\documentclass[final]{cvpr}

\usepackage{times}
\usepackage{epsfig}
\usepackage{graphicx}
\usepackage{amsmath}
\usepackage{amssymb}
\usepackage{multirow}
\usepackage{tabularx}
\usepackage{caption}
\usepackage{multicol}

\usepackage[pagebackref=true,breaklinks=true,colorlinks,bookmarks=false]{hyperref}

\def\cvprPaperID{4780} 
\def\confYear{CVPR 2021}

\begin{document}

\title{Few-Shot Model Adaptation for Customized Facial Landmark Detection, Segmentation, Stylization and Shadow Removal}

\author{Zhen Wei\\
EPFL\\
{\tt\small zhen.wei@hotmail.com}
\and
Bingkun Liu\\
HKUST\\
\and
Weinong Wang $\,\,\,\,\,\,\,\,\,$ Yu-Wing Tai\\
Kuaishou Technology
}

\twocolumn[{%
	\renewcommand\twocolumn[1][]{#1}%
	\maketitle
	\vspace{-8mm}
	\begin{center}
		\centering
		\includegraphics[width=1\linewidth]{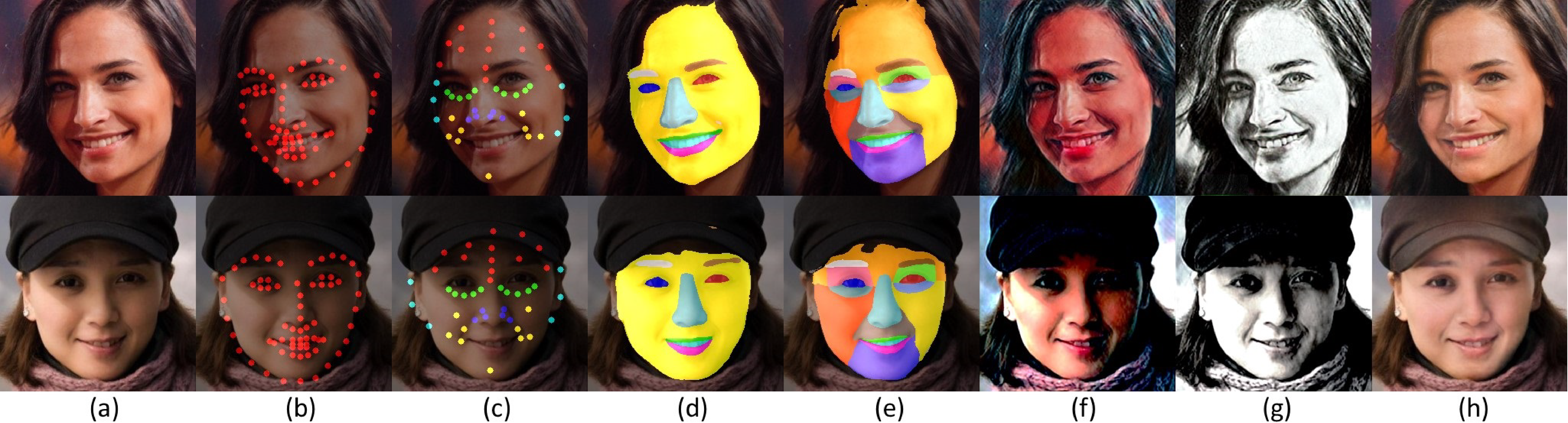}\vspace{-0.05in}
		\captionof{figure}{The FSMA's predictions on the web images (a) over different facial image applications, including (b) standard 68-point landmark detection, (c) customized landmark detection, (d) standard 9-class face segmentation, (e) customized face segmentation, (f)(g) customized artistic stylization and (h) facial shadow removal.
		From (b) to (h), each model is trained with \textbf{\textit{150, 20, 50, 50}} annotated images, \textbf{\textit{15, 15}} example images and \textbf{\textit{100}} images without shadow.
		The FSMA models corresponding to the tasks (b), (d) and (h) only utilize \textbf{\textit{4\%-5\%}} manual annotations comparing to the fully supervised methods.}
	\end{center}%
}]

\newcommand\blfootnotetex[1]{%
  \begingroup
  \renewcommand\thefootnote{}\footnotetext{#1}%
  \addtocounter{footnote}{-1}%
  \endgroup
}

\blfootnotetex{This work was done when the first two authors were student interns at Kuaishou Technology.}

\begin{abstract}
   Despite excellent progress has been made, the performance of deep learning based algorithms still heavily rely on specific datasets, which are difficult to extend due to labor-intensive labeling. Moreover, because of the advancement of new applications, initial definition of data annotations might not always meet the requirements of new functionalities. Thus, there is always a great demand in customized data annotations. To address the above issues, we propose the Few-Shot Model Adaptation (FSMA) framework and demonstrate its potential on several important tasks on Faces. The FSMA first acquires robust facial image embeddings by training an adversarial auto-encoder using large-scale unlabeled data. Then the model is equipped with feature adaptation and fusion layers, and adapts to the target task efficiently using a minimal amount of annotated images. The FSMA framework is prominent in its versatility across a wide range of facial image applications. The FSMA achieves state-of-the-art few-shot landmark detection performance and it offers satisfying solutions for few-shot face segmentation, stylization and facial shadow removal tasks for the first time. \vspace{-0.1in}
\end{abstract}

\begin{figure*}[t]
	\begin{center}
		\includegraphics[width=1\linewidth]{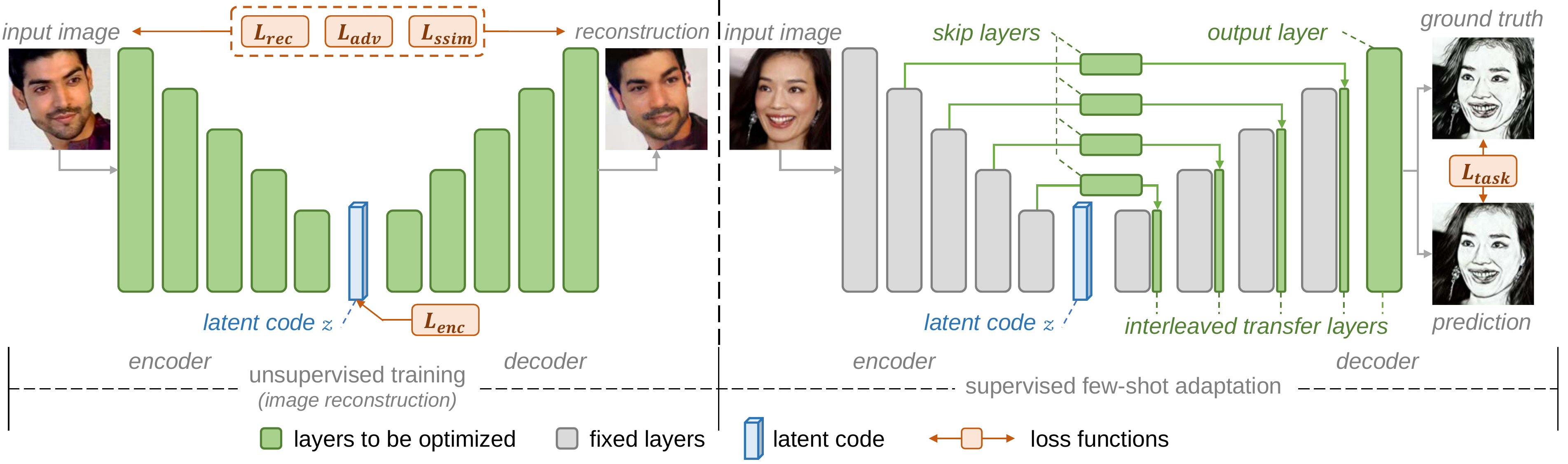}
	\end{center}
	\vspace{-6mm}
	\caption{
		The training pipeline and model structure of the FSMA framework.
		In the first unsupervised training stage, an adversarial auto-encoder is optimized for the image reconstruction task with a great amount of unlabeled facial images.
		To adapt to a customized task, the auto-encoder's backbone is fixed and the newly initialized interleaved transfer layers, skip layers and output layer are optimized.
	}
	\label{fig:framework}
	\vspace{-4mm}
\end{figure*}

\section{Introduction}
The advances achieved by deep learning models for facial landmark detection, segmentation, editing and enhancement tasks have made many new applications. 
However, the process of devising various deep learning applications is always accompanied by the challenges from data. 
On the one hand, training models on a large-scale public dataset with fixed annotation formats cannot fulfill the highly customized and fast-changing needs. 
For example, a landmark detection network trained on the 300W \cite{300W} dataset generates the standard 68-point landmarks, which are unable to model the forehead and face skin for advance editing. 
A segmentation model trained on Helen \cite{helen} dataset cannot parse out chin or cheeks for face analysis at a finer granularity. 
On the other hand, though it is feasible to annotate new datasets at the cost of much time and labor, the conventional fully supervised learning paradigm is inapt in practice when facing more diverse, flexible and even customized demands, hindering researchers from coming up with more inspiring and novel applications for fast development iterations. 
A better solution is enabling a model to obtain a satisfying performance with only a minimal number of customized annotation, namely the few-shot model adaptation.

Several recent works have attempted the few-shot model adaptation for facial images. 
They leverage either unsupervised learning or meta-learning techniques to handle the facial landmark detection with limited data \cite{3FacRec}, the video face reenactment with unseen identity \cite{intro_review_reenactment} and the face anti-spoofing for unknown attacks \cite{intro_review_antispoofing}. Nevertheless, these models are specifically designed for a single task. 
To our best knowledge, there is no a versatile few-shot adaptation method that can be easily extended to face analysis, editing and enhancement tasks within a unified framework.

In this paper, we proposed the Few-Shot Model Adaptation (FSMA) framework which aims at data-efficient model adaptation as well as the versatility across a wide range of applications. The training pipeline and the model's structure are shown in Figure \ref{fig:framework}.
The framework first exploits the compact feature embedding via large-scale unsupervised learning, in which an adversarial auto-encoder \cite{aae} is trained on the image reconstruction task.
The high-fidelity reconstruction ensures the auto-encoder to capture the high-level `face knowledge' \cite{face_knowledge}.
Then the auto-encoder is further trained on supervised tasks with a small number of annotated images.
So far, the similar idea \cite{3FacRec} has achieved progress in predicting the mid-level semantic information, e.g., facial landmarks.
However, due to the bottleneck-like structure and lack of internal feature resolution, the vanilla auto-encoder hinders itself from adapting to more subtle analysis, editing and enhancement tasks, like face semantic segmentation, stylization and shadow removal. 
To this end, we introduce additional skip layers between the encoder and decoder during the supervised adaptation stage.
The skip layers not only prevent the auto-encoder from cheating during the unsupervised learning but also realize the multi-scale feature fusion as well as the recovery of feature details.
As verified in the experiments (Table \ref{tab:skip_layer_ablation_lm}\ref{tab:skip_layer_ablation_seg} and Figure \ref{fig:res_skip_layer}), the introduced skip layers are critical components to FSMA's versatility on various applications. 

Extensive experiments are conducted on both the small customized datasets as well as the standard datasets under few-shot setting. 
To our best knowledge, the FSMA is the first solution to handle few-shot face semantic segmentation, few-shot facial image stylization and few-shot facial image shadow removal problems.
Both qualitative and quantitative results demonstrate the FSMA's superiority over existing task-specific models and versatility across different tasks.
On few-shot facial landmark detection task, the FSMA achieves state-of-the-art performance. 

	

\section{Related Works}
Representative works in few-shot methodology and each application area are reviewed in this section.

\subsection{Semi-Supervised Learning}
The FSMA framework is essentially a semi-supervised learning pipeline that follows the `unsupervised pretrain, supervised fine-tune' paradigm.
One series of related literature focus on the unsupervised pretraining stage, including the pseudo labeling methods \cite{related_2_pseudolabel,related_2_pseudolabel_1} and the self-supervised methods \cite{related_2_selfsup_1,related_2_selfsup_2,related_2_selfsup_3}.
Another series of works emphasized more on the supervised adaptation stage.
Task-specific efforts for a single applications include \cite{related_1_image_translation} for image translation, \cite{3FacRec} for facial landmark detection, \cite{related_1_segmentation} for image segmentation, etc.
Besides, \cite{simclrv2} proposed an extra distillation stage using unlabeled data after the supervised fine-tuning for further improvement.
Instead of working on a particular task, the FSMA framework is designed for a wide range of facial image applications and has superior versatility than other task-specific models.

\subsection{Applications on Facial Images}
The validation of FSMA's capacity is conducted on facial image analysis, editing and enhancement applications.
In detail, the few-shot problems in landmark detection, segmentation, stylization and shadow removal are discussed.

\vspace{1mm}
\noindent\textbf{Facial landmark detection.}
In this paper, we focus on the models that work with limited data.
Semi-supervised models such as \cite{Kingma2014} adopted partially and weakly annotated data. 
Solutions to generate pseudo label include landmark perturbation \cite{Lv2016}, multi-views captured from 3D face model \cite{Zhu2016}, image style translation \cite{sa} and GAN generation \cite{GAN_ref}.
The other attempts include multi-task learning \cite{Ranjian2016, Zhang2014}, `teacher-student' framework for training quality improvement \cite{TS3} and large-scale unsupervised learning \cite{3FacRec}.
Similar to \cite{3FacRec}, the FSMA framework also leverages large-scale unlabeled data, but is more expandable to various tasks.

\vspace{1mm}
\noindent\textbf{Face segmentation.}
Most previous works are studied under the fully supervised setting.
Studies on network architecture worked on hierarchical model \cite{Luo2012,aaai20_face_parsing}, image pyramids \cite{icnn}, multi-task learning \cite{cvpr15liusifei}, cascade model, CNN-RNN \cite{cnn_rnn_seg} hybrid network and receptive field \cite{adaptiveRF,tip_seg}.
\cite{tip_seg} also devised a new loss function and semi-supervised distillation.
\cite{tanh_warp} proposed a warping method to better segment hairs.
Different from these works, the FSMA is the first attempt to solve the few-shot face segmentation problem.

\vspace{1mm}
\noindent\textbf{Facial image stylization.}
On general images, \cite{Gatys2016} firstly introduced deep learning models. 
\cite{CycleGAN2017} developed a framework with unpaired images. 
\cite{PerceptualLosses} devised the Perceptual Loss for better generation quality. 
Other works have contributed to improve the result's visual quality \cite{Li2019, Xin2017, Artsiom2018, Yulun2019}, inference speed \cite{Li2016, Dmitry2016, CQF2017}, diversities \cite{Vincent2017, Dongdong2017, Yijun2017, Xun2017, Yijun20172} and controllablity \cite{Jing2017, Pierre2017, Gatys2017}.
As the application on facial images, several works addressed specific tasks, such as makeup/de-makeup \cite{Qiao2019,Wai2007,Ying2017}, photo-sketch translation \cite{Mingrui2017,Peng2020}, face aging \cite{face_aging_1,face_aging_2}, etc.
Comparing to previous works, the FSMA framework focuses on the few-shot model adaptation problem on the artistic stylization.

\vspace{1mm}
\noindent\textbf{Facial shadow removal.}
Most recent related works concentrated on the network designs, including using novel GAN \cite{Wang_2018,Cun_2020} and RNN \cite{Hu_2019,Ding_2019} structures. 
Besides, \cite{face_shadow_removal} manually selected a few thousand images without shadow and trained its model on synthesized shadowed images.
The FSMA framework aims at training with minimal data and alleviate the labor cost when collecting images for training.

\section{The FSMA Framework }
The goal of the FSMA framework is to reduce the manpower spent on data annotation and collection during the development of a customized application for facial images.
By leveraging large-scale unlabeled images and then a tiny number of annotated data, the FSMA enables a deep learning model to achieve satisfying performance on a new task.

The FSMA's training pipeline has two major stages: the unsupervised learning and the supervised adaptation stage.
Figure \ref{fig:framework} illustrates the pipeline and the model's structure.

\subsection{Exploiting Feature via Unsupervised Learning}
In the unsupervised learning stage, an adversarial auto-encoder is trained.
Following the framework proposed in \cite{aae,3FacRec}, the overall optimization objective comprises four loss functions.
An $L_1$ pixel-wise image reconstruction loss $L_{rec}$ and an adversarial loss $L_{adv}$ that penalize the difference between input and reconstructed image are used to guarantee the high-fidelity reconstruction ability.
An encoding feature loss $L_{enc}$ is applied to the latent vector $z$ to ensure its smoothness and continuousness in the feature space.
We also maximize the SSIM \cite{ssim} between input and output image as a regularization term $L_{ssim}$ to maintain the face structural information.
The final objective for this stage is a weighted sum of the loss functions in mention, which is
\[L_{unsup} = \lambda_1 L_{rec} + \lambda_2 L_{adv} + \lambda_3 L_{enc} + \lambda_4 L_{ssim}\,.\]
The loss weights are selected empirically where $\lambda_1=\lambda_2=\lambda_3=1.0$, $\lambda_4=60$.

\subsection{Supervised Few-Shot Adaptation}
\subsubsection{Modifications on Network's Structure}
The supervised few-shot adaptation learning aims to extract task-specific features from the acquired robust face representations, and facilitate the expected functionality with the extracted information.
Therefore, the parameters of the encoder and decoder are fixed.
This operation also prevents the model from overfiting on the small-scale annotated data.
To adapt the model to a particular application, interleaved transfer layers \cite{3FacRec} are added into the decoder.
Each interleaved transfer layer is a $3\times3$ convolution layer inserted before every upsampling operation.
Besides, the original output layer is replaced with a newly initialized one with different output channel number to fit the targeted task.

\subsubsection{Adding Skip Layers}
In practice, we find that the model is not capable enough to transfer to a wide range of tasks if only equipped with the interleaved transfer layers.
Due to its high compactness, the latent code $z$ only contains high-level information.
The decoder is not able to recover the detailed features that authentically represent the original facial image.
This factor always leads to inferior performance when the targeted task requires detailed features, e.g. for face segmentation, stylization and shadow removal.

\begin{figure}[t]
    \vspace{-2mm}
	\begin{center}
		\includegraphics[width=0.95\linewidth]{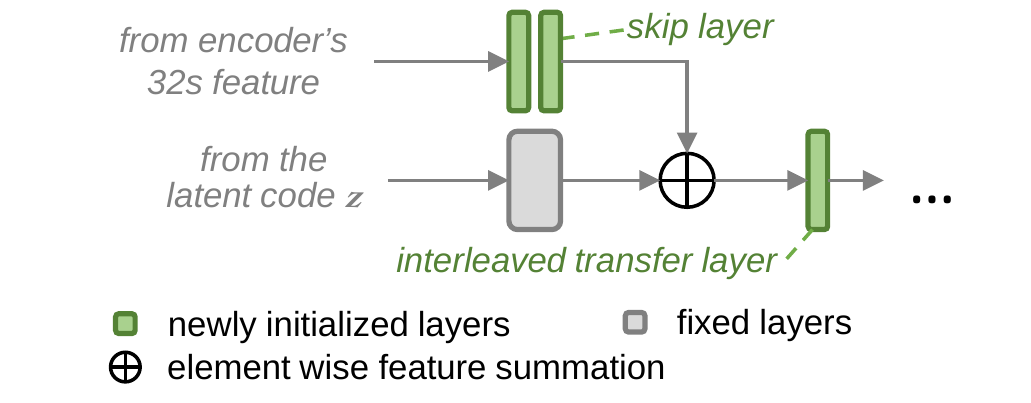}
	\end{center}
	\vspace{-5mm}
	\caption{
		The detailed structure of skip layers.
		This figure takes the feature with 32 downsampling rate (32s) as an example.
		The skip layer processes feature from the encoder and sum its output to the input for interleaved transfer layer.
	}
	\label{fig:skiplayer}
	\vspace{-4mm}
\end{figure}

To solve this problem, we introduce feature connections between the corresponding feature scales in encoder and decoder and add newly initialized adaptation layers in the connections.
We name these additional layers as the `skip layers'.
A skip layer consists of two $3\times3$ convolutional layers with nonlinear activations and feature normalizations.
It takes the feature from the encoder as input and its output is added to the input of the interleaved transfer layer.
An illustration of its detailed structure is shown in Figure \ref{fig:skiplayer}.

For conventional auto-encoders, the skip layers are intrinsically prohibited during its training.
It is because the short-circuited feature connections break the `compression and de-compression' model structure and degrade the learned feature quality.
However, the FSMA framework adopts skip layers during its supervised adaptation stage with its backbone fixed, which avoids the failure of unsupervised learning while offering multi-scale features as needed in the targeted task.

The skip layers are the critical components in achieving the FSMA's versatility over different types of facial image applications.
The absence of some or all skip layers not only leads to less robustness to data scarcity, but also directly causes malfunctioning in editing and enhancement tasks.

To give a concise representation, we use binary digits to denote the existence of skip layers.
For example, `00000' stands for a model without skip layer while `11111' means all the five feature scales have a skip layer. 
`11000' shows the presences for features with 32 and 16 times down-sampling.
\label{sec:digit}
Note that the `0000'/`00000' model is identical to \cite{3FacRec}, which serves as a baseline in the ablation studies.

\subsubsection{Optimization Objective}
The optimization objective $L_{task}$ in the supervised model adaptation stage is task-specific and is commonly used in fully supervised methods.
Its details are elaborated in Section \ref{sec:app} for each application.
More detailed experiment settings are also contained in the supplementary file.

\section{Applications}
\label{sec:app}
Without loss of generality, we validate FSMA's adaptation performance and task versatility on face analysis, editing and enhancement applications.
Specific tasks include facial landmark detection and segmentation for face analysis, facial image stylization for face editing and facial shadow removal for face enhancement.

Besides, we also introduce customized data annotation formats for these tasks.
The customized data not only serves for FSMA's validation but also complements existing common datasets and remedies their failures.

\begin{figure}[t]
    \vspace{-2mm}
	\begin{center}
		\includegraphics[width=1\linewidth]{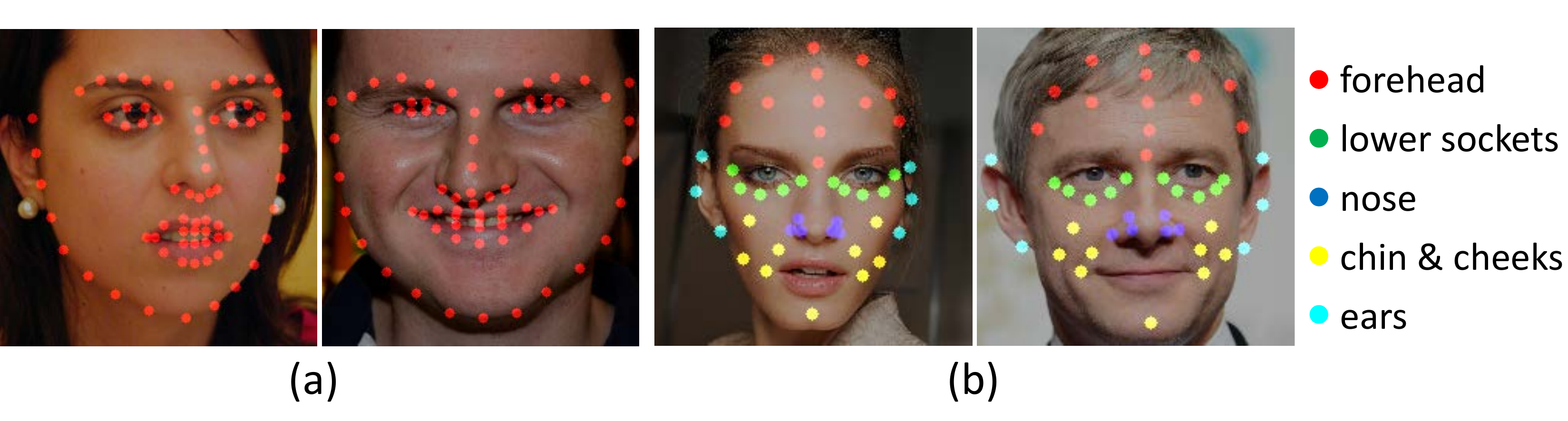}
	\end{center}
	\vspace{-8mm}
	\caption{
		A demonstration of (a) standard 68-point landmark format from 300W dataset and (b) the customized 44-point complementary annotation.
	}
	\label{fig:lm_annotations}
	\vspace{-5mm}
\end{figure}

\subsection{Facial Landmarks}
Facial landmark detection locates semantic key points and describes the basic facial structure.
Therefore, it is a fundamental component in many applications.

\textbf{Customized data.}
The public datasets adopt predefined standard landmark formats to ensure a model can match the same local area in different faces.
For example, the 
300W dataset uses a 68-point format where the keypoints distribute along the face edge and facial components. However, the drawbacks are also obvious. 
The foreheads are empty, making the landmarks unable to bound the whole facial area.
Existing formats focus much more on facial components than the face skin where it is less textured but still varies on geometry.
 
To overcome these issues, we define additional 44 landmarks as complementary to the 68-point format.
As shown in Figure \ref{fig:lm_annotations}, the new keypoints mainly focus on the face skin, among which 13 points cover the forehead and hairline, ten points on the lower eye sockets to circle around the eyes' with original points on eyebrows, six points locate nostrils and alae while others define chin and cheeks.
We annotate only 20 images collected from the 300W dataset.
The annotation workload costs less than an hour for an engineer.

\textbf{Learning objective.}
During the adaptation, the model generates a heatmap for each landmark with half of the input image's resolution.
Therefore, there are four feature scales in the decoder and the binary skip layer indicator  has only four digits.
The $L_{task}$ is an $L_2$ distance between the ground truth and predicted heatmaps.
Experiments are conducted on both the 300W and customized data.


\subsection{Face Semantic Segmentation}
A face segmentation model is to assign each pixel with a class, which is a combination of high-level semantic analysis and low-level detailed structure prediction.

\textbf{Customized data.}
Since this task requires pixel-wise ground truth, the annotation is highly time-consuming.
Academic researches have been based on small-scale datasets (e.g. LFW \cite{LFW}, Helen \cite{helen}) over a long time.
Even the latest dataset \cite{aaai20_face_parsing} contains much less images than those for other facial image tasks.
Except for the shortage of labeled images, the annotation format is also over-simplified.
Current datasets divide the whole face into the face skin and facial components.
Similar to facial landmarks, the public segmentation datasets emphasize more on facial components than the variances inside face skin.

\begin{figure}[t]
\vspace{-4mm}
	\begin{center}
		\includegraphics[width=1.01\linewidth]{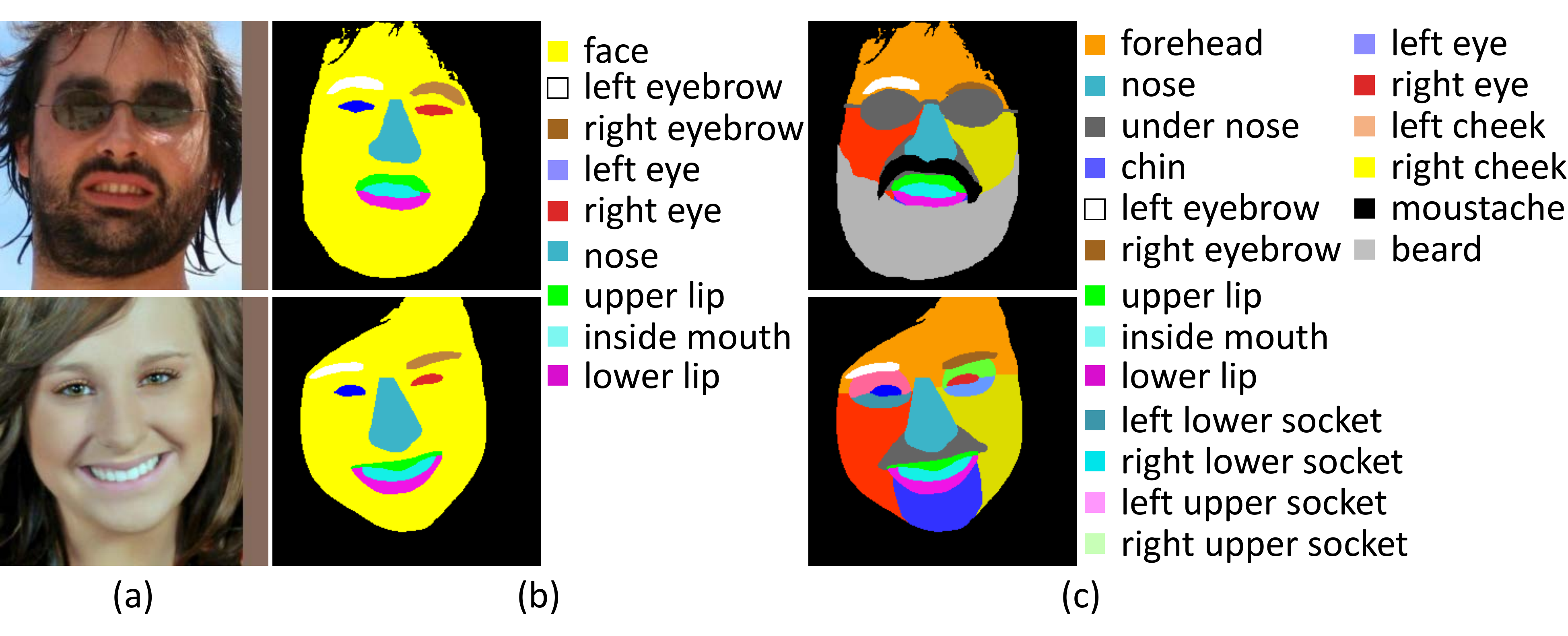}
	\end{center}
	\vspace{-8mm}
	\caption{
		A demonstration of (b) Helen dataset's annotation and (c) the customized complementary annotation, which correspond to the original images (a).
	}
	\label{fig:seg_annotations}
	\vspace{-4mm}
\end{figure}

Here we propose a format that complements the Helen's annotation.
Specifically, the original face skin is divided into the forehead, left and right cheeks, chin, under nose area and eye sockets.
The eye sockets are further partitioned into left/right and upper/lower sockets, which are four parts in total.
The masks for common accessories like glasses, beard and mustache are also labeled.
Few examples are given in Figure \ref{fig:seg_annotations}.
The more fine-grained definition enables a model to distinguish various facial textures and geometries, and improve the performance of applications such as 3D face reconstruction and animation.

\textbf{Learning objective.}
During adaptation training, the model output predicted masks that have the same size as input images.
The cross-entropy loss is used as $L_{task}$. 


\subsection{Facial Image Stylization}

\begin{figure}[t]
	\begin{center}
		\includegraphics[width=1\linewidth]{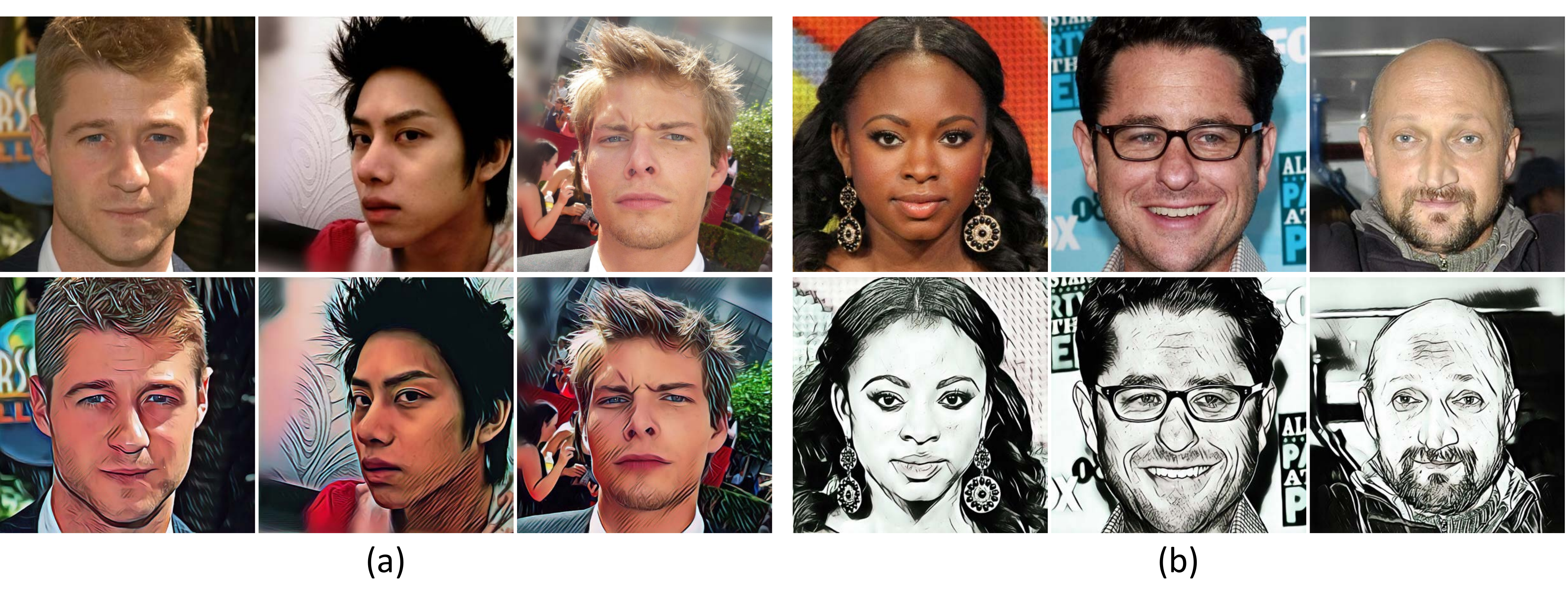}
	\end{center}
	\vspace{-8mm}
	\caption{
		Examples of the customized data for few-shot stylization training.
		Each column contains an original image and its stylized counterpart.
		There are two different visual styles (a) and (b).
	}
	\label{fig:style_annotations}
	\vspace{-5mm}
\end{figure}

Image stylization translates an image to a different visual domain and brings global visual changes.
This technique has been widely applied in many applications, such as face cartoonization and artistic stylization.

\textbf{Customized Data.}
Being an image editing task, training a stylization model always lacks enough direct supervision.
On the one hand, it costs a long time for an artist to process or paint the paired illustrations on a large scale.
On the other hand, the industrial demand for high-quality stylization emerges and changes rapidly.
In this case, building a conventional fully supervised pipeline to develop a stylization model is not feasible.

We verify FSMA's capability for the few-shot style adaptation by training the model with a few pairwise data.
To this end, we randomly collect a few images from the CelebA-HQ \cite{celeba,celebahq} dataset and generate two different art styles with a commercial application.
This process is equivalent to the artist's manual image processing.
Several pairwise data are shown in Figure \ref{fig:style_annotations}.
The chosen styles contain significant color and texture changes.

\textbf{Learning objective.}
During the supervised adaptation training, the $L_{rec}$ and $L_{adv}$ used in the unsupervised learning stage are kept as the task-specific loss $L_{task}$.

\subsection{Facial Shadow Removal}
\begin{figure}[t]
    \vspace{-3mm}
	\begin{center}
		\includegraphics[width=0.9\linewidth]{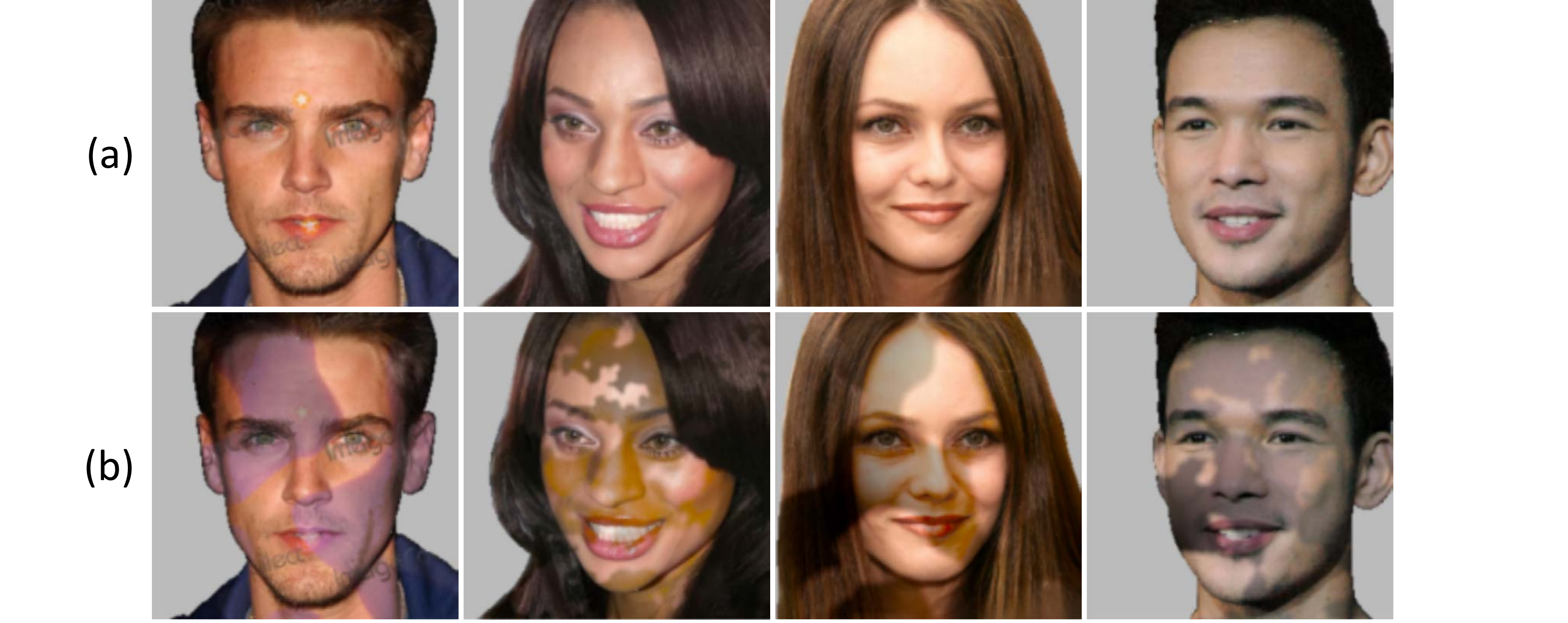}
	\end{center}
	\vspace{-6mm}
	\caption{
		Examples of the synthesized facial shadow.
		(a): the face areas.
		(b): images with random shadows.
	}
	\label{fig:shadow_annotations}
	\vspace{-5mm}
\end{figure}
Unlike facial image stylization that performs global transformation over an image, facial image enhancement usually makes local visual changes.
Facial shadow removal is one of the enhancement applications which aims to compensate for the brightness within the shadow on the face caused by foreign objects.
This technique improves an image's aesthetic quality, especially when the photographer is limited by the environmental lighting conditions.

\textbf{Few-shot data collection.}
A recent solution \cite{face_shadow_removal} is to generate synthetic shadows on facial images.
In this case, the human labor is involved in choosing the images without foreign shadows, or the `clean images'.
To synthesize training data, binary masks are first randomly generated using the Perlin noise or certain object silhouettes.
Then the pixel intensities inside the masks are decayed to imitate illumination changes.
Several synthesized images are shown in Figure \ref{fig:shadow_annotations}.
The \cite{face_shadow_removal} used 5000 clean images in total.
The data collection in this amount is demanding and it would be more difficult for other customized needs.
As for the FSMA framework, only a minimal number of clean images are collected (less to 1\%).

\textbf{Learning objective.}
The $L_{task}$ used here is the same as that for the stylization task.
To focus on FSMA's adaptation ability, other task-specific techniques in \cite{face_shadow_removal} are not used.


\begin{figure*}[t]
    \vspace{-3mm}
	\begin{center}
		\includegraphics[width=1\linewidth]{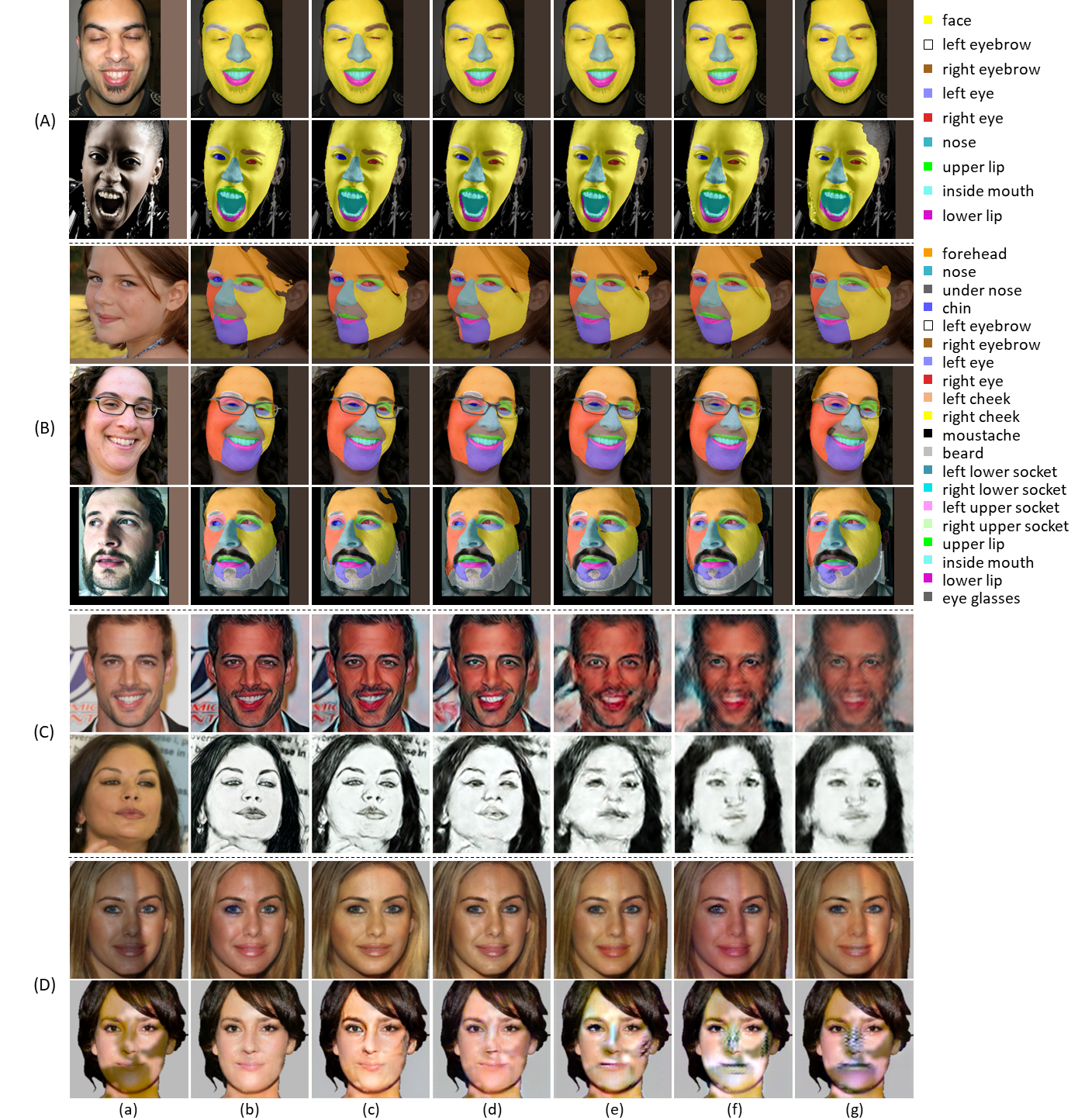}
	\end{center}
	\vspace{-6mm}
	\caption{
		Qualitative comparisons of the models for (A),(B) face segmentation, (C) image stylization and (D) shadow removal  with skip layer setting of (b) 11111, (c) 11110, (d) 11100, (e) 11000, (f) 10000 and (g) 00000.
		(a) are input images from (A)(B) Helen validation set, (C) CelebA-HQ dataset and (D) validation images with synthesized shadows.
	}
	\label{fig:res_skip_layer}
	\vspace{-4mm}
\end{figure*}

\begin{table}[htbp]
    \vspace{-2mm}
	\centering
	\caption{The ablation study on the skip layer settings w.r.t. the model's landmark detection accuracy on 300W validation set.
	The skip layer setting is defined in \ref{sec:digit}.
    Each result contains the inside/outlier/all landmark accuracies.
}
\resizebox{.35\textwidth}{!}{
	\begin{tabular}{c|ccc}
		\hline
		\textbf{skip layer settings} & \multicolumn{3}{c}{\textbf{NME(\%)}} \\
		\hline
		0000  & 3.65  & 6.40  & 4.31  \\
		1000  & 3.64  & 6.31  & 4.29 \\
		1100  & 3.61  & 6.24  & 4.15 \\
		1110  & 3.42  & 6.17  & 4.13 \\
		1111  & \textbf{3.12}  & \textbf{6.14}  & \textbf{3.88}  \\
		\hline
	\end{tabular}%
}
	\label{tab:skip_layer_ablation_lm}%
\end{table}%

\begin{table}[t]
	\centering
	\caption{
	    The ablation study on the skip layer     settings w.r.t. the model's segmentation   accuracy on Helen validation set.
	    The skip layer setting is defined in \ref{sec:digit}.
		Skip layers and feature fusions are critical to model's performances.
	}
\resizebox{.46\textwidth}{!}{
	\begin{tabular}{c|c|c|c|c|c|c}
		\hline
		\textbf{skip layer} & \multicolumn{6}{c}{\textbf{F1 scores (\%)}}  \\
		\cline{2-7}    \textbf{settings} & \textbf{face} & \textbf{eyebrows} & \textbf{eyes} & \textbf{nose} & \textbf{mouth} & \textbf{overall} \\
		\hline
		00000 & 92.10  & 73.55  & 79.11  & 90.71  & 87.62  & 85.23  \\
		10000 & 92.83  & 75.15  & 80.00  & 92.36  & 88.25  & 86.60  \\
		11000 & 93.69  & 76.51  & 83.44  & 92.35  & 89.89  & 87.78  \\
		11100 & 94.23  & 78.33  & 85.81  & 93.22  & 91.28  & 89.21  \\
		11110 & 94.70  & \textbf{81.31}  & \textbf{87.91}  & 93.28  & 92.11  & 90.09  \\
		11111 & \textbf{94.72}  & 80.74  & 87.81  & \textbf{93.67}  & \textbf{92.47}  & \textbf{90.32}  \\
		\hline
	\end{tabular}%
}
	\label{tab:skip_layer_ablation_seg}%
\end{table}%

\begin{figure}[t]
	\begin{center}
		\includegraphics[width=1\linewidth]{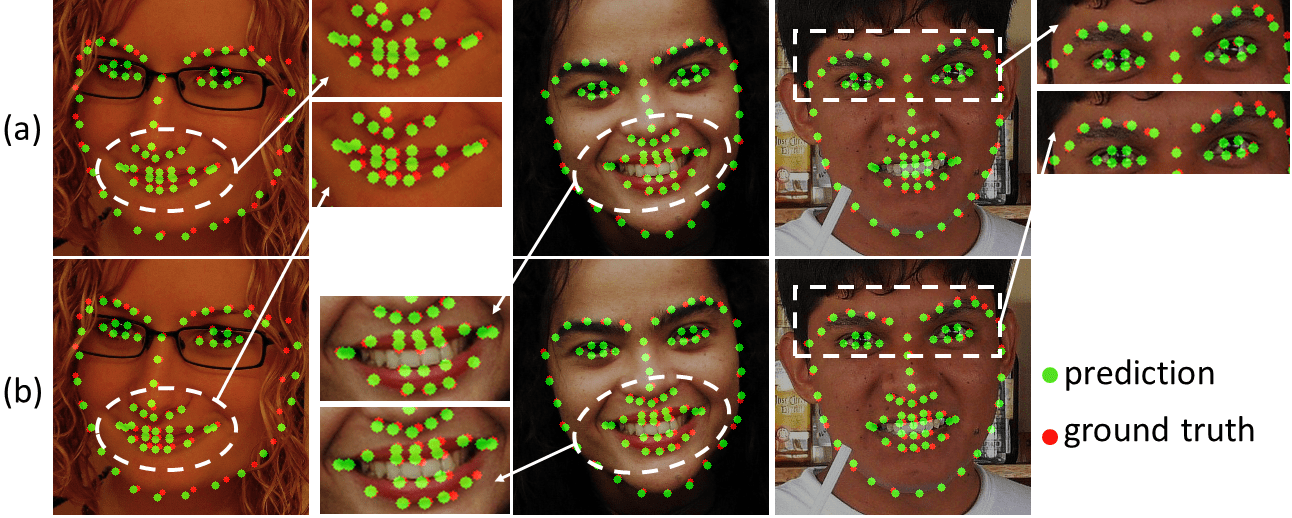}
	\end{center}
	\vspace{-6mm}
	\caption{
		Qualitative comparisons between the models with skip layer setting (a) 1111 and (b) 0000.
		Major differences are highlighted.
		Best viewed with zoom in.
	}
	\label{fig:res_lm_skip_layer}
	\vspace{-4mm}
\end{figure}

\section{Ablation Studies \& Results}
Detailed experimental settings including dataset description, hyper-parameter selection and evaluation metrics for the unsupervised learning and adaptation are elaborated in the supplementary files.

\subsection{The Skip Layers and FSMA's Versatility.}
The skip layers are the main factor that contributes to FSMA's task versatility.
They improve the model's prediction accuracy on face analysis tasks while, on face editing and enhancement tasks, the skip layers determine the model's success and failure.
From a quantitative view, Table \ref{tab:skip_layer_ablation_lm}, \ref{tab:skip_layer_ablation_seg} show that more skip layers promote landmark detection and segmentation consistently in all columns, which indicates that the improvement for feature quality is comprehensive.
As for qualitative comparisons, the predicted landmarks around the mouth and eyes shown in Figure \ref{fig:res_lm_skip_layer} locate closer to the ground truth.
Figure \ref{fig:res_skip_layer}(A)(B) show that the skip layers not only help to parse more detailed structures, like the glass frame, but also can identify whether the person's eyes are closed or not.
The results of stylization models in Figure \ref{fig:res_skip_layer}(C) indicate that the model where skip layers are absent fails to produce stylized detailed texture and facial structures.
The output images are degraded noticeably in columns (f) and (g).
Furthermore, the models shown in Figure \ref{fig:res_skip_layer}(D) with fewer and no skip layers fail to capture the shadow areas, not to mention to remove the shadow.

\begin{table*}[htbp]
	\centering
	\caption{The quantitative comparison with state-of-the-art facial landmark detection models under the few-shot training setting on 300W validation set.
	Results are reported under the NME(\%) metric.
	The three values in a single result stand for the performances of inside/outlier/all landmarks.
	The training set size is proportional to 300W's training set.
	FSMA's binary digits indicate the presence of skip layers as defined in \ref{sec:digit}.
}
    \vspace{-4mm}
    \resizebox{.80\textwidth}{!}{
	\begin{tabular}{l|ccc|ccc|ccc|ccc}
		\hline
		\multirow{2}[4]{*}{\textbf{models}} & \multicolumn{12}{c}{\textbf{training set size}} \\
		\cline{2-13}          & \multicolumn{3}{c|}{100\%} & \multicolumn{3}{c|}{20\%} & \multicolumn{3}{c|}{10\%} & \multicolumn{3}{c}{5\%} \\
		\hline
		RCN+\cite{rcn}  & 4.20  & 7.78  & 4.90  & -     & 9.56  & 5.88  & -     & 10.35  & 6.32  & -     & 15.54  & 7.22  \\
		SA\cite{sa}    & 3.21  & \textbf{4.98}  & \textbf{3.46}  & 3.85  & -     & -     & 4.27  & -     & -     & 6.32  & -     & - \\
		TS$^3$\cite{TS3} & \textbf{2.91}  & 5.90  & 3.49  & 4.31  & 7.97  & 5.03  & 4.67  & 9.26  & 5.64  & -     & -     & - \\
		3FabRec\cite{3FacRec} & 3.36  & 5.74  & 3.82  & 3.76  & \textbf{6.53}  & 4.31  & 3.88  & \textbf{6.88}  & 4.47  & 4.22  & \textbf{6.95}  & 4.75  \\
		\hline
		FSMA  &       &       &       &       &       &       &       &       &       &       &       &  \\
		$\,\,$-0000 & 3.65  & 6.40  & 4.31  & 3.63  & 7.08  & 4.49  & 4.05  & 7.36  & 4.87  & 4.13  & 7.78  & 5.04 \\
		$\,\,$-1111 & 3.12  & 6.14  & 3.88  & \textbf{3.23}  & 6.87  & \textbf{4.14}  & \textbf{3.59}  & 7.01  & \textbf{4.45}  & \textbf{3.85}  & 7.30  & \textbf{4.71}  \\
		\hline
	\end{tabular}%
	}
	\vspace{-2mm}
	\label{tab:sota_lm}%
\end{table*}%

\begin{table*}[t]
	\centering
	\caption{The quantitative comparison with state-of-the-art face segmentation models on Helen validation set.
	* means the model utilizes extra annotations.}
	\vspace{-2mm}
    \resizebox{.82\textwidth}{!}{
	\begin{tabular}{l|c|c|c|c|c|c|c|c}
		\hline
		\multirow{2}{*}{\textbf{models}} & \textbf{image} & \multirow{2}{4em}{\textbf{backbone}}  & \multicolumn{6}{c}{\textbf{F1 scores (\%)}} \\
		\cline{4-9}          & \textbf{resolution} &       & \textbf{face}  & \textbf{eyebrows} & \textbf{eyes}  & \textbf{nose}  & \textbf{mouth} & \textbf{overall} \\
		\hline
		MO-GC\cite{cvpr15liusifei} & 250$\times$250   & other & 91.0  & 71.3  & 76.8  & 90.9  & 84.1  & 84.7 \\
		iCNN\cite{icnn}  & 256$\times$256   & other & -     & 81.3  & 87.4  & \textbf{95.0}  & \textbf{92.6}  & 87.3  \\
		CNN-RNN\cite{cnn_rnn_seg} & 256$\times$256   & other  & 92.1  & 77.0  & 86.8  & 93.0  & 89.1  & 88.6  \\
		Adaptive RF\cite{adaptiveRF} & 256$\times$256   & VGG16 & 91.48  & 78.61  & 84.66  & 93.65  & 91.48  & 90.21  \\
		FSMA  & 256$\times$256   & Res18  & \textbf{94.72}  & \textbf{80.74}  & \textbf{87.81}  & 93.67  & 92.47  & \textbf{90.32} \\
		\hline
		BSPNet+LaPa*\cite{aaai20_face_parsing} & 473$\times$473   & Res101     & 95.1  & 81.9  & 87.8  & 94.7  & 93.8  & 91.4 \\
		AFSRTS*\cite{tip_seg} & 512$\times$512   & VGG16     & 95.62  & 83.88 & 89.92  & 94.73  & 94.31  & 92.04  \\
		Tanh-Warp*\cite{tanh_warp} & 512$\times$512   & Res18     & 95.3  & 85.9  & 89.7  & 95.6  & 95.2  & 93.1  \\
		\hline
	\end{tabular}%
}
	\label{tab:sota_seg}%
	\vspace{-4mm}
\end{table*}%

\begin{figure}[t]
    \vspace{-2mm}
	\begin{center}
		\includegraphics[width=1\linewidth]{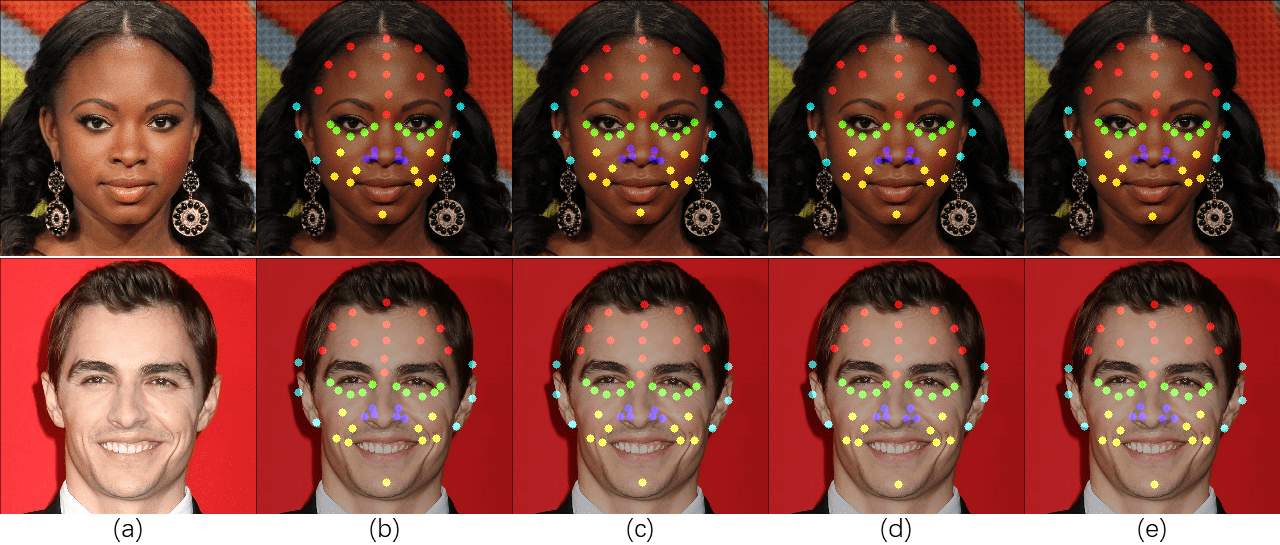}
	\end{center}
	\vspace{-6mm}
	\caption{
		Qualitative comparisons of landmark detection models trained on (b) \textbf{\textit{20}}, (c) \textbf{\textit{10}}, (d) \textbf{\textit{5}} and (e) \textbf{\textit{1}} images.
	}
	\label{fig:res_customized_lm_shot}
	\vspace{-2mm}
\end{figure}

\begin{figure}[t]
	\begin{center}
		\includegraphics[width=1\linewidth]{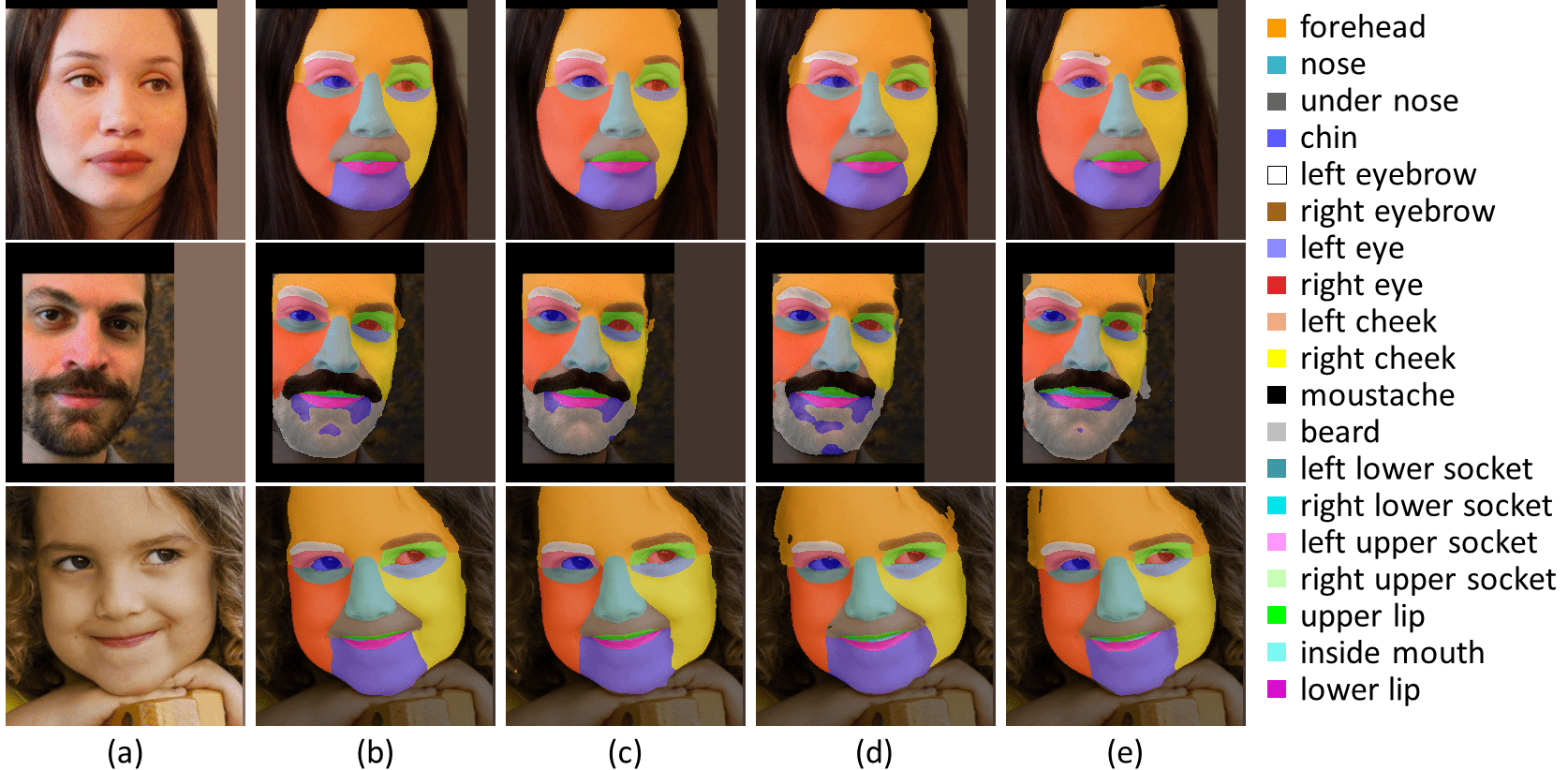}
	\end{center}
	\vspace{-6mm}
	\caption{
		Qualitative comparisons of face segmentation models trained on (b) \textbf{\textit{100}}, (c) \textbf{\textit{50}}, (d) \textbf{\textit{25}} and (e) \textbf{\textit{10}} images.
	}
	\label{fig:res_customized_seg_shot}
	\vspace{-2mm}
\end{figure}

\begin{figure}[t]
	\begin{center}
		\includegraphics[width=1\linewidth]{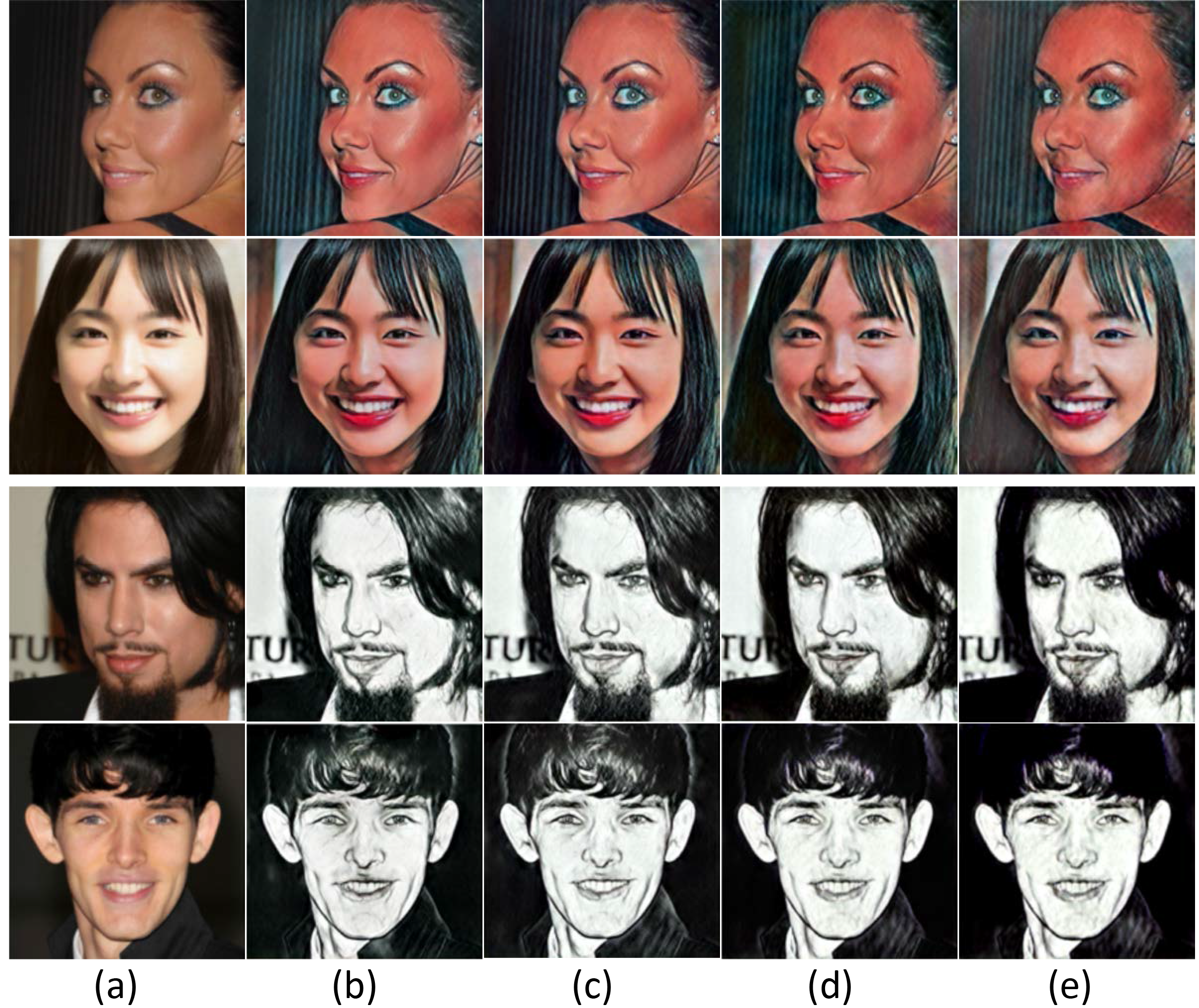}
	\end{center}
	\vspace{-6mm}
	\caption{
		Qualitative comparisons of the FSMA stylization models with (b) \textbf{\textit{50}}, (c) \textbf{\textit{25}}, (d) \textbf{\textit{15}} and (e) \textbf{\textit{10}} `clean images'.
	}
	\label{fig:res_style_shot}
	\vspace{-4mm}
\end{figure}

\subsection{Effect of the Amount of Training Data}
\begin{figure}[t]
\begin{center}
	\includegraphics[width=1\linewidth]{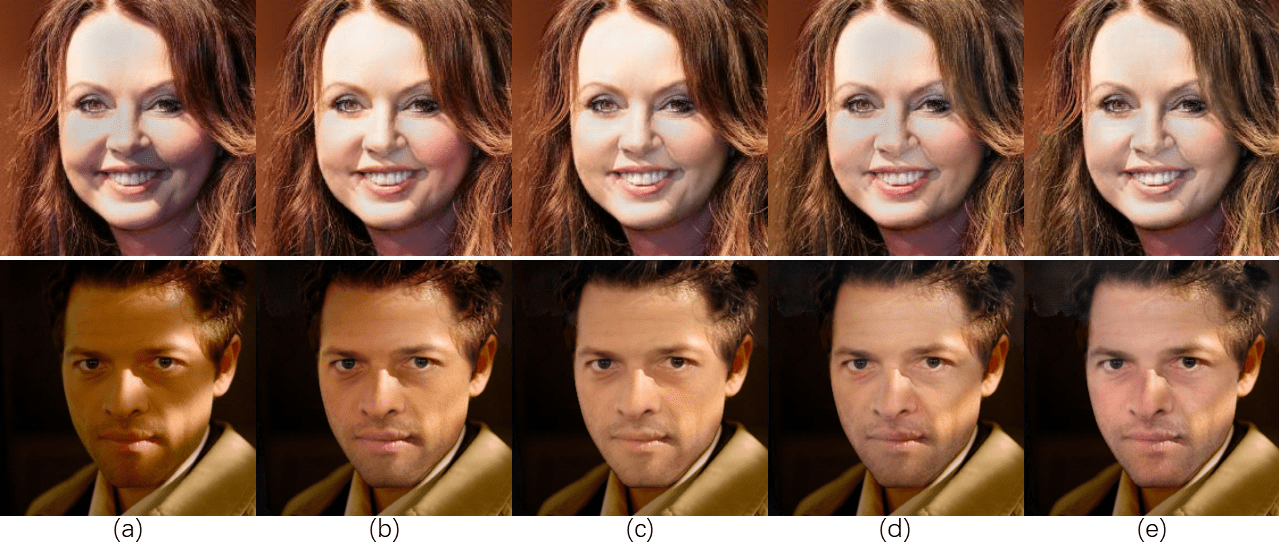}
\end{center}
\vspace{-4mm}
\caption{
	Qualitative comparisons of facial shadow removal models with (b) \textbf{\textit{2500}}, (c) \textbf{\textit{500}}, (d) \textbf{\textit{100}} and (e) \textbf{\textit{50}} `clean images' manually selected for training.
	(a) are images with natural shadow.
	For a comparison, \cite{face_shadow_removal} used \textbf{\textit{5000}} `clean images' under the same training protocol.
}
\label{fig:res_shadow_shot}
\end{figure}

Figure \ref{fig:res_customized_lm_shot}-\ref{fig:res_shadow_shot} demonstrate the qualitative comparisons among the FSMA models given various numbers of training images.
For all the tasks except stylization, the models generate visually satisfying results with less to 1\% training supervision compared to previous fully supervised models.
Meanwhile, as presented in Figure \ref{fig:res_style_shot}, the models correspond to column (e) uses only 10 pairwise training data while maintaining most expected visual details in its outputs.
From the quantitative aspect, Table \ref{tab:sota_lm} shows that the FSMA model has the least performance drop compared with state-of-the-art few-shot models.
In table \ref{tab:image_number_ablation_seg}, the FSMA model is much more robust to data scarcity than the ResNet18 \cite{resnet} based UNet \cite{unet} model, which is identical to the `11111' model's structure but is initialized with ImageNet \cite{imagenet} pretrained weights.
The FSMA segmentation model also outperforms many fully supervised state-of-the-art methods with only 5\% training data on Helen dataset.

\subsection{Comparison with State-of-the-Arts}
We compare the FSMA facial landmark detection with recent models under the few-shot setting and face segmentation model with state-of-the-art methods under the fully supervised setting.
Table \ref{tab:sota_lm} demonstrates the FSMA's superiority for landmarks on facial components and the overall landmarks given less training images.
Note that the `0000' FSMA model is theoretically identical to \cite{3FacRec}.
For the segmentation model compared in in Table \ref{tab:sota_seg}, the FSMA model with the same input image resolution and supervision outperforms other models even with a less powerful backbone.

\begin{table}[t]
	\centering
	\caption{
		The ablation study on the number of training images w.r.t. the model's F1 score on Helen val set.
		The training set size in percentage is proportional to Helen's training set.
	    The skip layer setting is defined in \ref{sec:digit}.
	    The UNet model has the same structure as FSMA's `11111' model but is initialized with the ResNet-18 pretrained on ImageNet.
	}

\resizebox{.48\textwidth}{!}{
	\begin{tabular}{c|c|c|c|c|c|c}
		\hline
		\multirow{2}{3em}{\textbf{models}} & \multicolumn{6}{c}{\textbf{training set size}} \\
		\cline{2-7}          & 100\% & 25\%  & 5\%   & 2.5\% & 0.5\% & 5 \\
		\hline
		UNet\cite{unet} & 89.53  & 88.66  & 87.09  & 82.71  & 75.49  & 71.19   \\
		\hline
		FSMA  &       &       &       &       &       &  \\
		$\,\,\,$-00000 & 85.23  & 84.99  & 84.28  & 82.77  & 78.72  & 75.21  \\
		$\,\,\,$-11111 & \textbf{90.32}  & \textbf{90.08}  & \textbf{89.10}  & \textbf{87.01}  & \textbf{84.53}  & \textbf{81.87}  \\
		\hline
	\end{tabular}%
}
	\label{tab:image_number_ablation_seg}%
	\vspace{-4mm}
\end{table}%

\begin{table*}[tbp]
	\centering
	\caption{Empirical comparisons on annotation workloads between the conventional fully supervised pipeline and the few-shot learning pipeline with FSMA.}
	\vspace{-2mm}
	\resizebox{.75\textwidth}{!}{
			\begin{tabular}{l|llll|cl}
				\hline
				& \multicolumn{4}{l|}{\textbf{man-hour/day for labeling 2k images}} & \multicolumn{2}{l}{\textbf{man-hour/day w/ FSMA}}\\
				\cline{2-7} \textbf{tasks}     & \multicolumn{1}{c}{first round } & \multicolumn{1}{c}{\multirow{2}[2]{*}{check}} & \multicolumn{1}{c}{\multirow{2}[2]{*}{correction}} & \multicolumn{1}{c|}{\multirow{2}[2]{*}{overall}} & \multicolumn{1}{c}{annotated} & overall\\
				& \multicolumn{1}{c}{annotation} &            &            &            & \multicolumn{1}{c}{images} & \\
				\hline
				lm.        & 100h/12.5d   & 2h$\;\;\,$/0.3d     & 2.8h$\;\;$/0.4d     & 108h/13.3d   & 100        & 6h$\;\;$/0.8d\\
				seg.       & 500h/42.5d & 3.2h/0.4d     & 16.8h/2.1d    & 522h/45.2d & 50         & 12h/1.5d \\
				\hline
			\end{tabular}%
	}
	\label{tab:annotation_worklad}
\end{table*}%

\begin{figure}[t]
\begin{center}
	\includegraphics[width=1\linewidth]{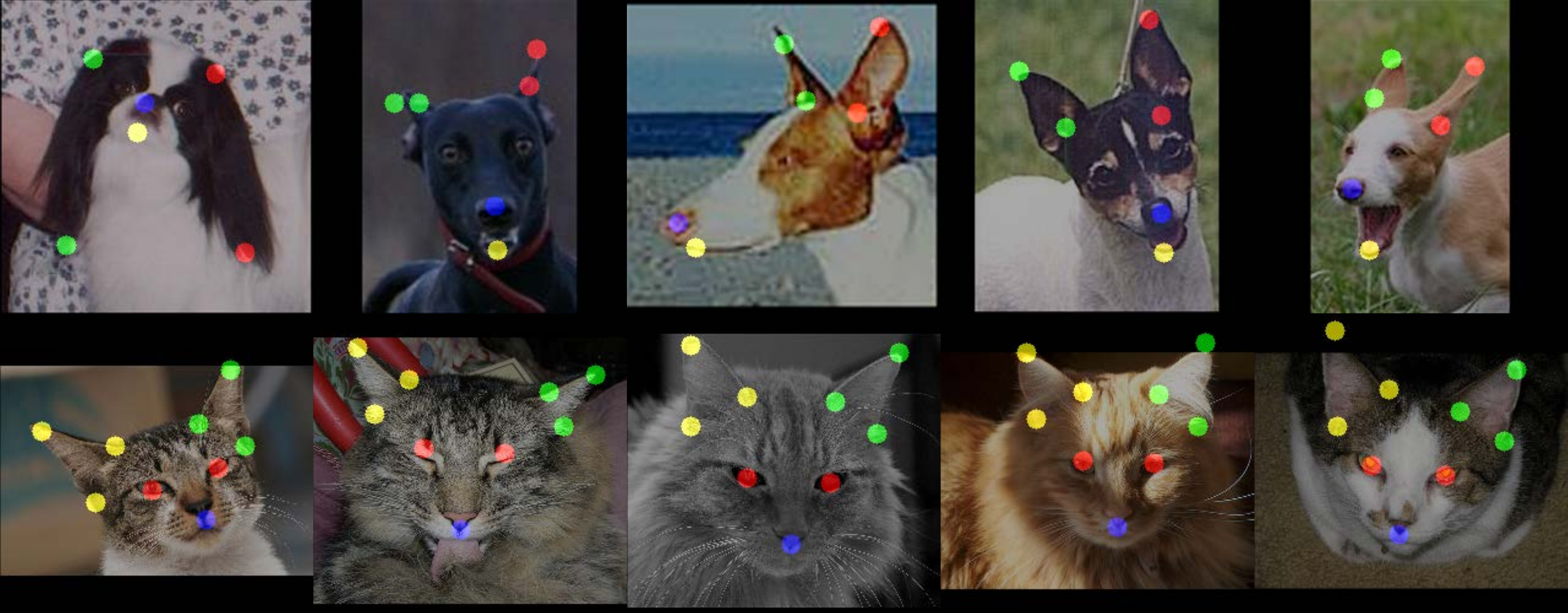}
\end{center}
\vspace{-2mm}
\caption{
		The customized landmark detection results of the FSMA-1111 model applied on the non-human facial images.
		15\% of StanfordExtra \cite{dog_dataset} and Cat-Head \cite{cat_dataset} training images are used for the few-shot adaptations.
}
\label{fig:cat_dog}
\end{figure}

\subsection{Reduction of Annotation Workloads}
In this paper, we emphasize on extending the few-shot model's functionalities to a wider range of customized facial image applications and producing practical values.
To demonstrate the significant advantages of FSMA in reducing the amount of annotation workloads, we conduct quantitative comparisons based on our empirical observations before and after deploying the FSMA.
We choose the landmark detection and segmentation tasks in this study since they have more predictable overall workloads.
We compare the man-hour/day costs of labelling 2k images following a conventional fully-supervised training pipeline as well as labelling a small number of data with the few-shot adaptation pipeline FSMA.
The workloads in both settings are minimum in order to achieve adequate model performances in developing customized applications for facial images.
As shown in Table \ref{tab:annotation_worklad}, the overall labor costs have drastically decreased with the usage of FSMA.

\subsection{Extendibility on Non-Human Facial Images}
FSMA's efficacy on non-human faces is also investigated.
We verify the detection ability for customized landmarks on cats' and dogs' faces using the StanfordExtra \cite{dog_dataset} and Cat-Head \cite{cat_dataset} dataset.
To process the data, the head regions are detected and cropped out, contributing to 12000 and 7008 images during the unsupervised pretraining for dogs and cats, respectively.
With more than 100 times less data for pretraining comparing to human facial images, the FSMA manages to obtain descent performances with 15\% of the original training sets for adaptation.
Figure \ref{fig:cat_dog} shows the examples.


\section{Conclusion}
In this paper, we have proposed a novel Few-Shot Model Adaptation (FSMA) framework and demonstrate its performance and versatility on various important computer vision tasks: facial landmark detection, face semantic segmentation, facial image stylization and facial shadow removal. These tasks requires both high-level semantic understanding and low-level detail reconstruction. In the unsupervised training stage, our adversarial auto-encoder is trained with millions of face images to capture face knowledge through reconstruction. In the model adaption stage, our design of skip layers, interleaved layers and output layers have provided a pathway to reconstruct the output of different tasks in high quality. Thus, enabling us to achieve the state-of-the-art performance in few-shot landmark detection, and satisfying good results for other tasks even with very limited data. We hope our FSMA framework can inspire and spark future research in few-shot learning applications.

{\small
\bibliographystyle{ieee_fullname}
\bibliography{egbib}
}

\setcounter{section}{0}

\onecolumn

\noindent\textbf{\Large {\centering Supplementary File}}
\vspace{4mm}

In this supplementary file, we include the additional experiment results to FSMA's robustness to data scarcity comparing to a fully supervised model and verify FSMA's cross dataset generalization ability.
In the second part of this file, we also explain the experimental settings that are not fully elaborated in the main paper.

\section{Additional Experiment Results}
\subsection{Qualitative Comparisons with the Fully Supervised Method}
In Table 3, 5 in the main paper, we compare the FSMA model with the other few-shot landmark methods and the UNet \cite{unet} segmentation model, which demonstrate that the FSMA models are more robust to data scarcities on these tasks.
The FSMA `0000'/`00000‘ settings, namely the 3FabRec \cite{3FacRec} model, are also compared in Table 1, 2, 3, 5 and Figure 8, 9.

\begin{figure*}[t]
    \vspace{-4mm}
	\begin{center}
		\includegraphics[width=0.71\linewidth]{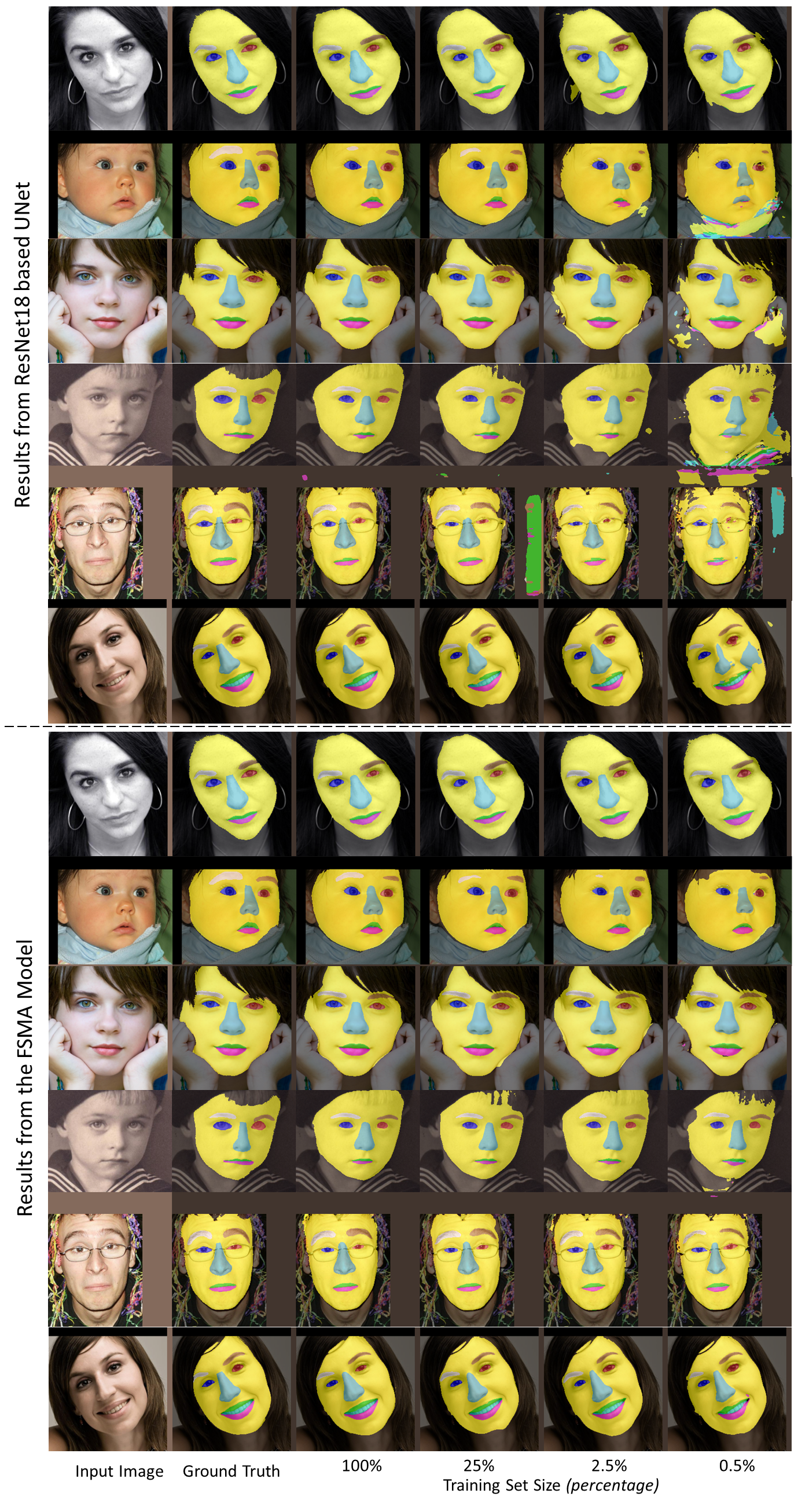}
	\end{center}
	\vspace{-4mm}
	\caption{
		The qualitative comparisons between the FSMA 9-class segmentation models and ResNet18 based UNet models on Helen's val set with different amount of training images.
	}
	\label{fig:supp_seg}
\end{figure*}

\begin{figure*}[t]
    \vspace{-4mm}
	\begin{center}
		\includegraphics[width=0.69\linewidth]{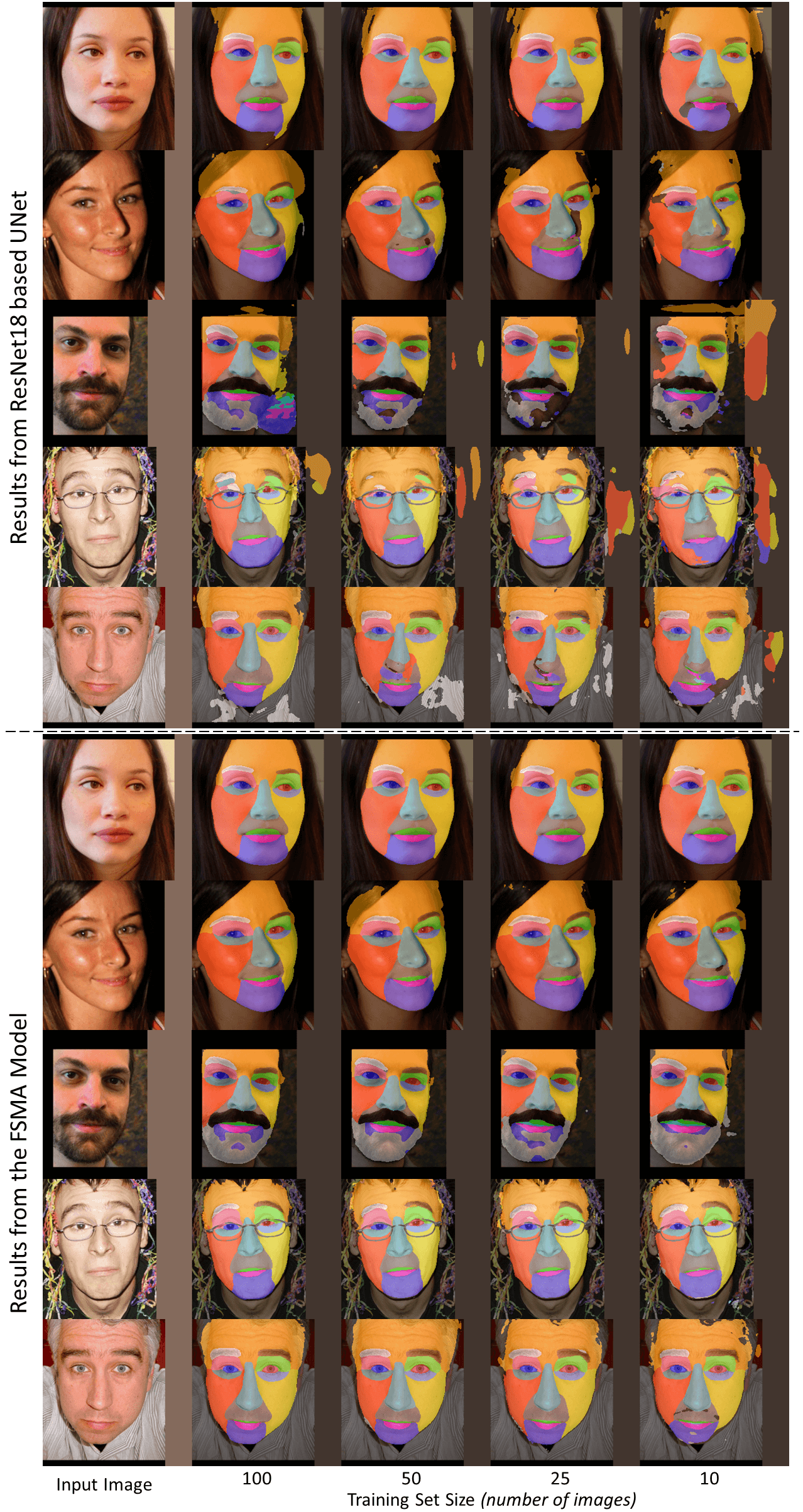}
	\end{center}
	\vspace{-4mm}
	\caption{
		The qualitative comparisons between the FSMA customized segmentation models and ResNet18 based UNet models on Helen's val set with different amount of training images.
	}
	\label{fig:supp_seg_customized}
\end{figure*}

\begin{figure*}[t]
    \vspace{-6mm}
	\begin{center}
		\includegraphics[width=0.63\linewidth]{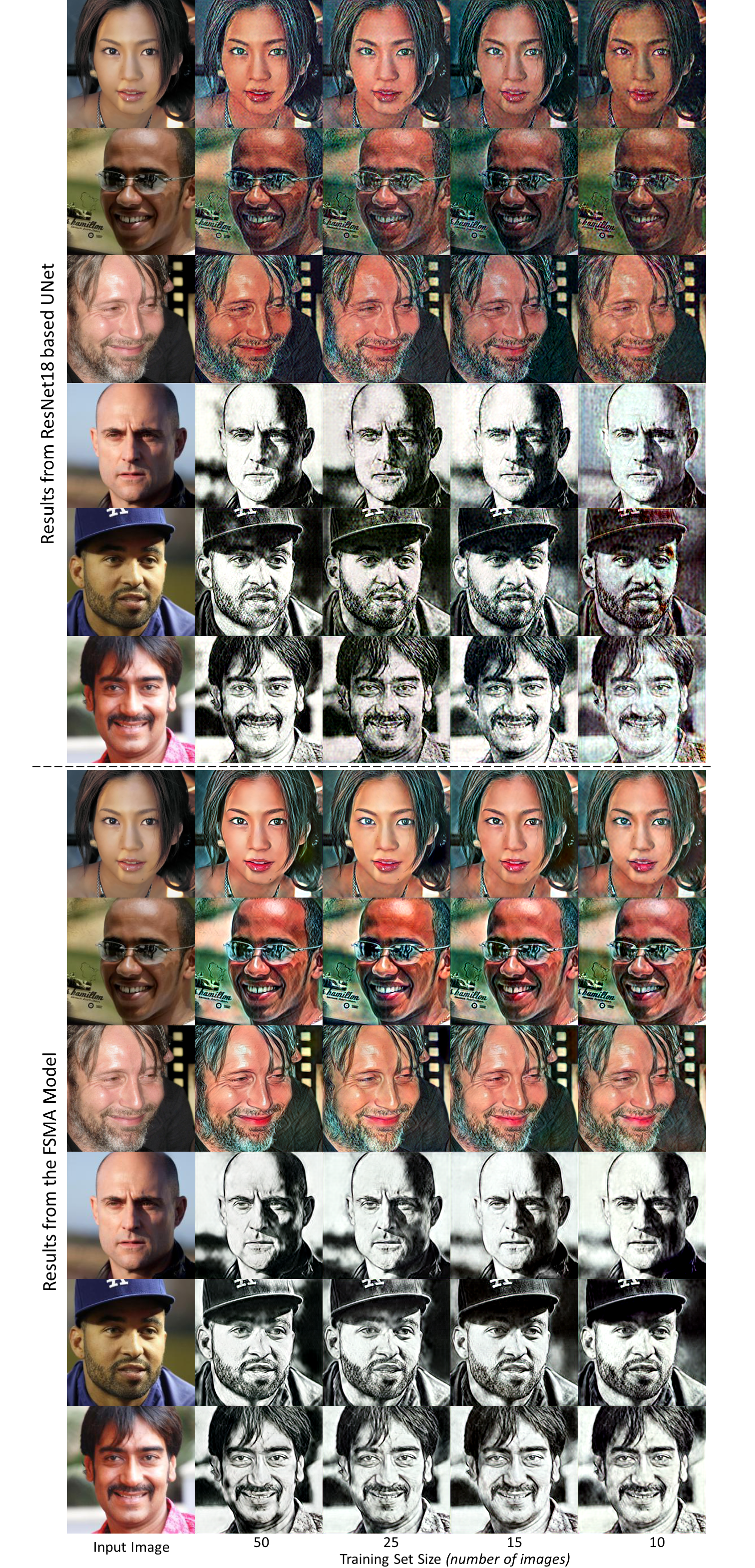}
	\end{center}
	\vspace{-4mm}
	\caption{
		The qualitative comparisons between the FSMA customized segmentation models and ResNet18 based UNet models with different amount of training images.
	}
	\label{fig:supp_style}
\end{figure*}

\begin{figure*}[t]
    \vspace{-6mm}
	\begin{center}
		\includegraphics[width=1.02\linewidth]{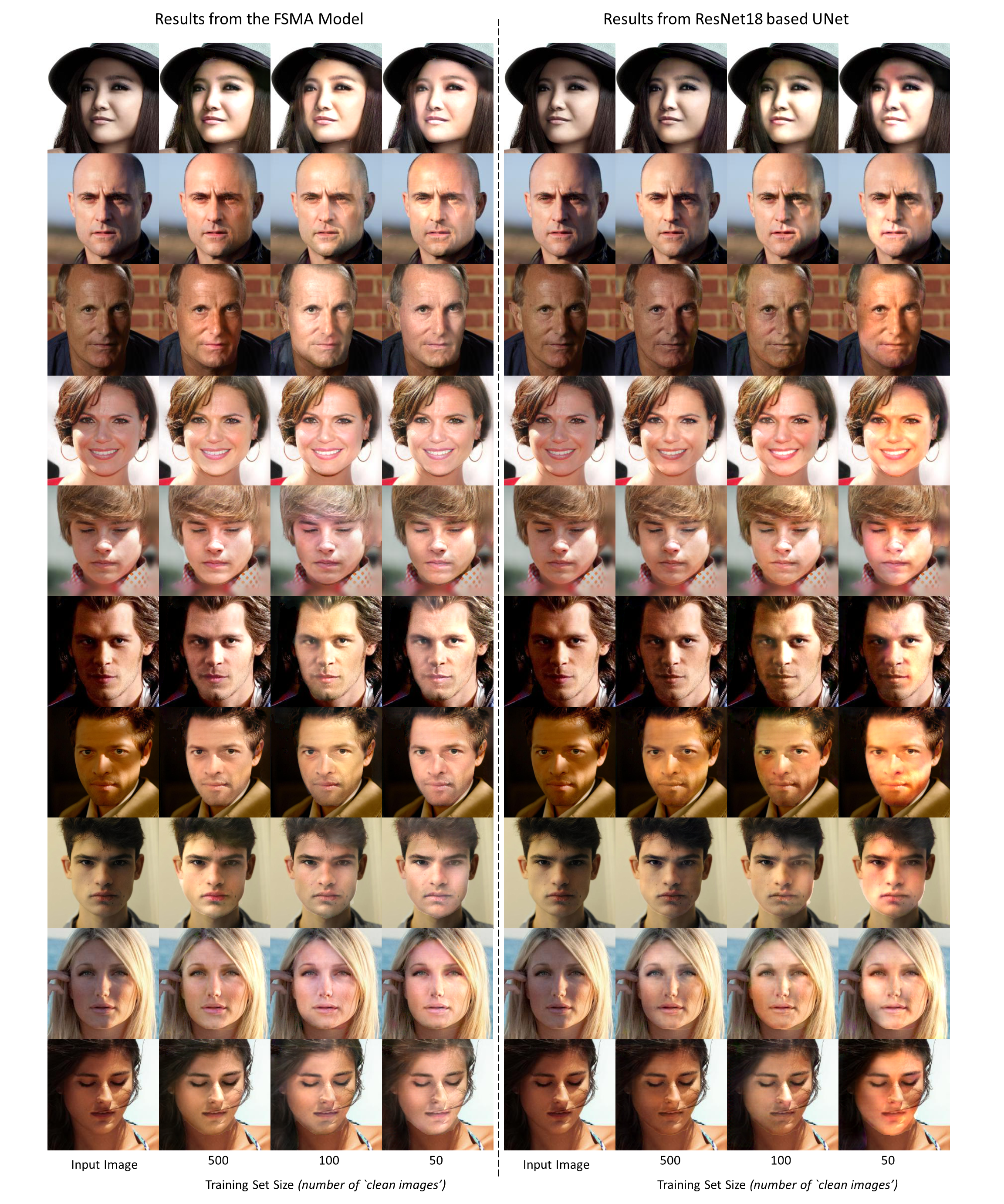}
	\end{center}
	\vspace{-4mm}
	\caption{
		The qualitative comparisons between the FSMA customized segmentation models and ResNet18 based UNet models with different amount of `clean images' for training.
	}
	\label{fig:supp_shadow}
\end{figure*}

In this section, we give more comparisons between the FSMA models and the ResNet18 based UNet model on segmentation, stylization  and shadow removal tasks.
By using the UNet model, we can ensure a fair comparison for few-shot adaptation capacity under the same number of parameters, training settings and network backbone.
In Figure \ref{fig:supp_seg}, \ref{fig:supp_seg_customized}, the FSMA segmentation models generalize much better than the UNet counterpart on smaller training set.
In Figure \ref{fig:supp_style}, \ref{fig:supp_shadow}, the FSMA stylization and shadow removal models generate less pixel noises and artifact.
Meanwhile, the FSMA shadow removal model is less prone to fail with minimal `clean images' used for training.
Note that we don't include the qualitative comparisons for landmark detection models since their differences are less visually perceptible.
The corresponding quantitative studies are listed in Table 3 in the main paper.

As a conclusion, the FSMA framework provides much few-shot adaptation ability for a model.

\begin{figure*}[t]
    \vspace{-6mm}
	\begin{center}
		\includegraphics[width=0.9\linewidth]{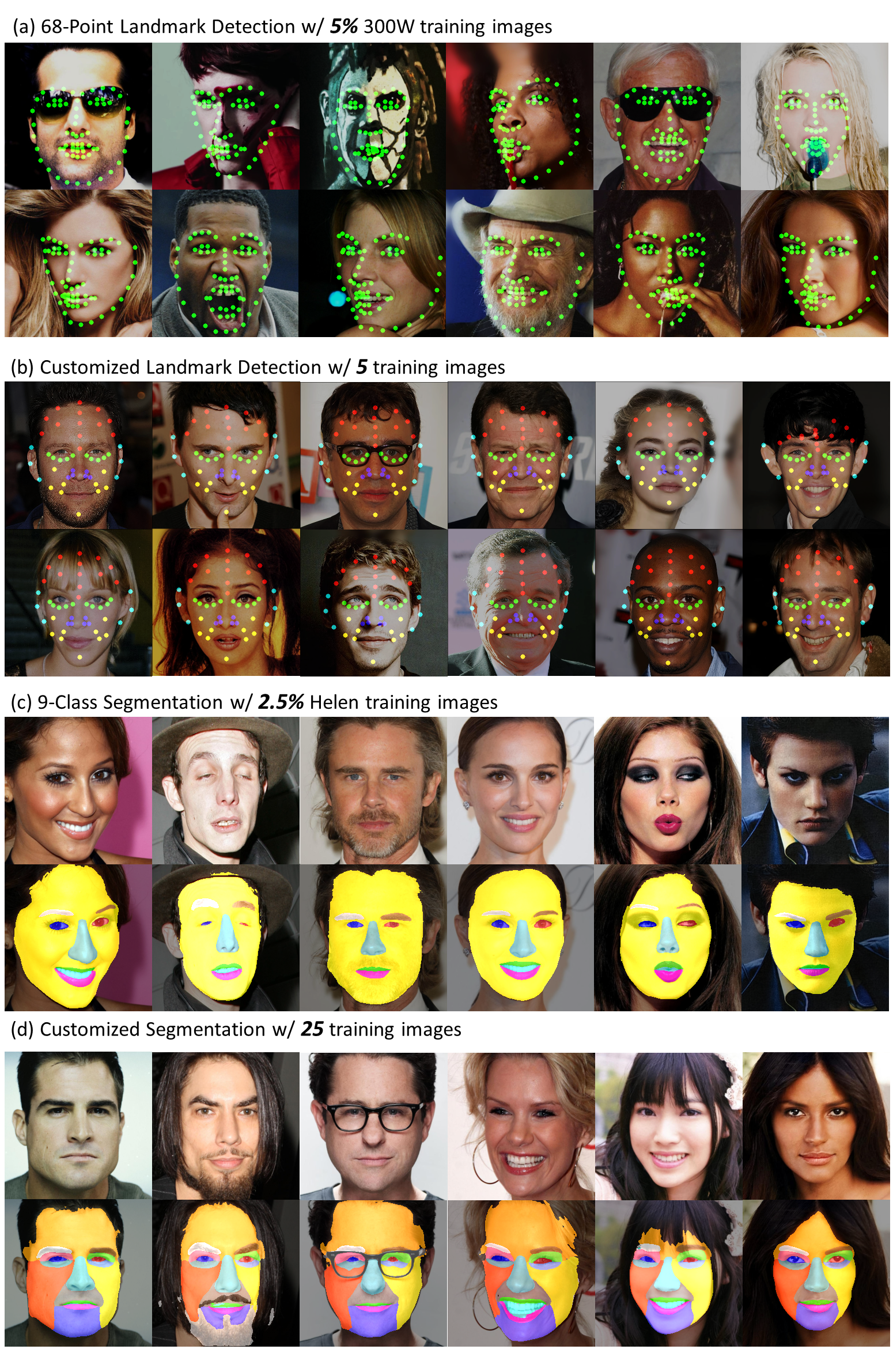}
	\end{center}
	\vspace{-4mm}
	\caption{
		Predictions from FSMA landmark detection models and segmentation models of images from the CelebA-HQ dataset.
		The results demonstrate FSMA's cross dataset generalization ability.
	}
	\label{fig:supp_res_1}
\end{figure*}

\begin{figure*}[t]
    \vspace{-6mm}
	\begin{center}
		\includegraphics[width=0.9\linewidth]{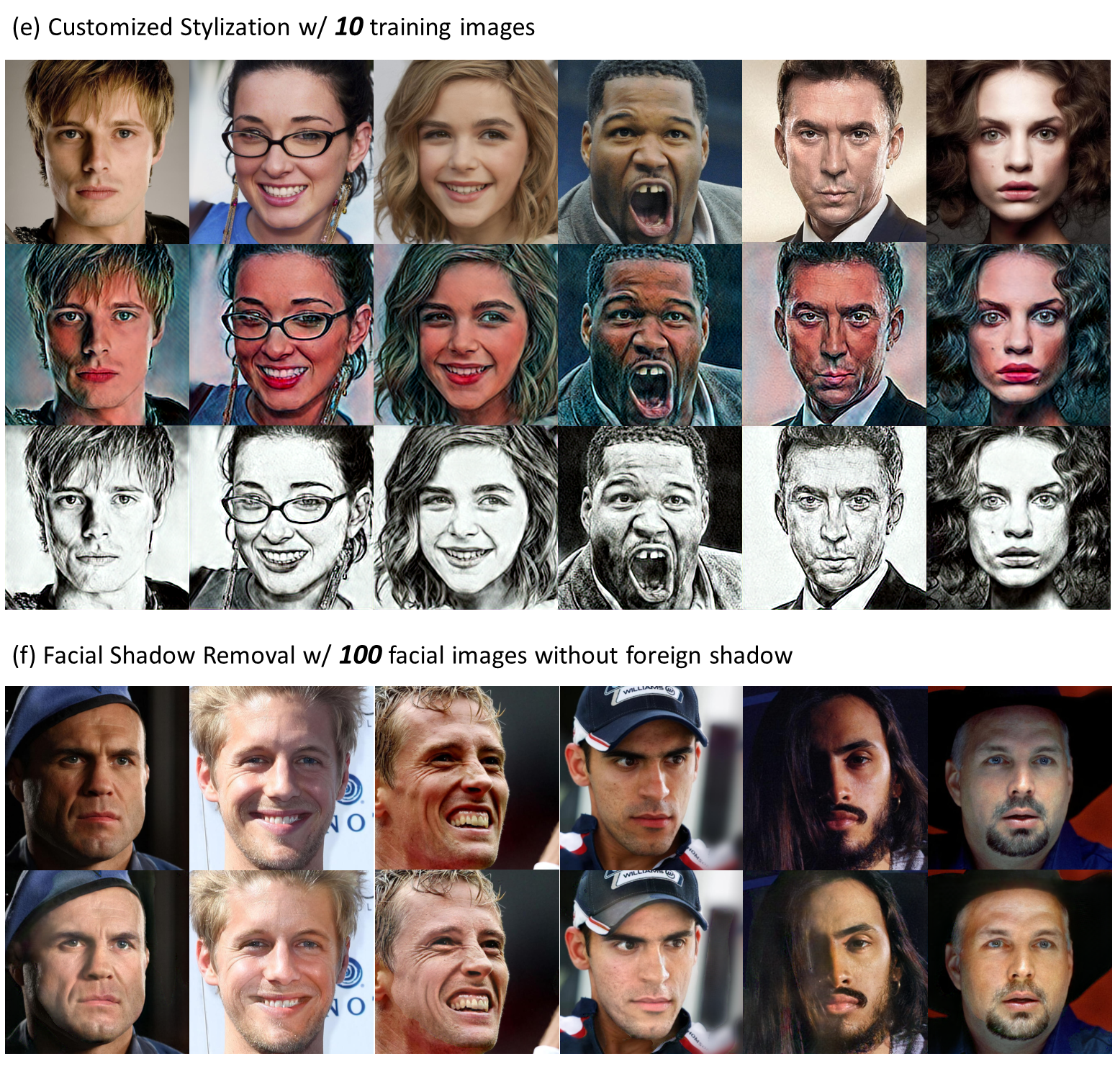}
	\end{center}
	\vspace{-4mm}
	\caption{
		Predictions from FSMA stylization models and shadow removal models of images from the CelebA-HQ dataset.
		The results demonstrate FSMA's cross dataset generalization ability.
	}
	\label{fig:supp_res_2}
\end{figure*}

\subsection{FSMA's Cross Dataset Generalization Ability}
In this section, we provide more prediction results of FSMA models on images randomly collected from the CelebA-HQ dataset with various conditions, which are shown in Figure \ref{fig:supp_res_1}, \ref{fig:supp_res_2}.
The sampled images follow the different distribution from FSMA models' training sets.

\section{Dataset Settings and Evaluation Metrics}
\subsection{Data for Unsupervised Learning}
During the large-scale unsupervised learning of the adversarial auto-encoder, we utilize the images collected by VGGFace2 \cite{vggface2} and AffectNet \cite{affectnet} datasets.
The VGGFace2 dataset contains 3.3 million facial images with various pose, age, illumination conditions.
The AffectNet dataset has 228k images, which especially captures a variety of facial expressions and thus provide diverse types of shapes of the face as well as facial components.
In practice, we remove the low-quality faces that may degrade the image reconstruction learning by discarding the images where the heights of faces are less than 100 pixels.
Therefore, the number of used images is 2.1 million in total.

\subsection{Data and Evaluation Metrics for Facial Image Applications}
\noindent\textbf{Facial Landmark Detection}

The 300W \cite{300W} dataset is used for the 68-point landmark detection task and the quantitative comparison with previous few-shot models.
The 300W dataset is assembled by \cite{300W} from the LFPW \cite{LFW}, AFW \cite{helen}, Helen \cite{helen_landmark}, XM2VTS \cite{xm2vtsdb} datasets.
All images are re-annotated with the 68-point landmark format.
With the common dataset split, there are 3,148 and 689 images in the training and testing set.
The latter one is further split into 554 images as the \textit{common} subset and 135 images as the \textit{challenge} subset.
The 300W dataset also has a private test set with 300 indoor and 300 outdoor images.
Under the few-shot setting, we use the whole dataset and randomly sample 20\%, 10\% and 5\% training images for the proposed FSMA model.

To build up the customized landmark detection dataset, we collect 20 images from 300W training set and annotate them with the format as described in Section 4.1.
We test the FSMA model's performances with 20, 10, 5 and 1 training image.
Note that the collected data are frontal images without neither heavy occlusions nor extreme poses.
The customized data are annotated with the LabelMe \cite{labelme2016} toolbox.

The performances of landmark detection models are evaluated under the normalized mean error (NME) metric.
The NME is the averaged distance between predicted and the ground truth landmarks that is divided by the distance of outer eye-corners as the `inter-ocular' normalization.

\begin{table*}[t]
  \centering
  \caption{The experimental settings of different applications during the supervised model adaptation stage.}
    \begin{tabular}{r|c|c|c|l|r}
    \hline
    \multicolumn{1}{c|}{\multirow{2}[2]{*}{\textbf{training data size}}} & \textbf{training} & \textbf{batch} & \textbf{learning} & \multicolumn{1}{c|}{\multirow{2}[2]{*}{\textbf{optimizer}}} & \multicolumn{1}{c}{\multirow{2}[2]{*}{\textbf{others}}} \\
          & \textbf{epochs} & \textbf{size} & \textbf{rate} &       &  \\
    \hline
    \multicolumn{6}{l}{\textbf{68-point landmark detection}} \\
    \hline
    100\% & 100   & \multirow{4}[2]{*}{20} & \multirow{4}[2]{*}{1 $\times 10^{-2}$} & \multicolumn{1}{l|}{Adam:} & \multicolumn{1}{l}{output size: 128 $\times$ 128,} \\
    20\%  & 500   &       &       & $\beta_1=0.9$, & \multicolumn{1}{l}{w/ data aug: } \\
    10\%  & 900   &       &       & \multicolumn{1}{l|}{$\beta_2=0.999$} & \multicolumn{1}{l}{random mirror, } \\
    5\%   & 900   &       &       &       & \multicolumn{1}{l}{random affine} \\
    \hline
    \multicolumn{6}{l}{\textbf{customized landmark detection}} \\
    \hline
    20    & \multirow{4}[4]{*}{1000} & 20    & \multirow{4}[4]{*}{$1\times10^{-2}$} & \multicolumn{1}{l|}{Adam: } & \multicolumn{1}{l}{w/o data aug} \\
    10    &       & 10    &       & $\beta_1=0.9$, &  \\
\cline{6-6}    5     &       & 5     &       & $\beta_2=0.999$, & \multicolumn{1}{l}{use IN} \\
    1     &       & 1     &       & \multicolumn{1}{l|}{w/ amsgrad} & \multicolumn{1}{l}{in skip layers} \\
    \hline
    \multicolumn{6}{l}{\textbf{9-class segmentation}} \\
    \hline
    100\% & 100   & \multirow{3}[2]{*}{50} & \multirow{4}[4]{*}{$1\times10^{-2}$} & \multicolumn{1}{l|}{Adam:} & \multicolumn{1}{l}{w/ data aug: } \\
    25\%  & 500   &       &       & $\beta_1=0.9$, & \multicolumn{1}{l}{random mirror, } \\
    2.5\% & 2000  &       &       & \multicolumn{1}{l|}{$\beta_2=0.999$} & \multicolumn{1}{l}{random affine} \\
\cline{3-3}    0.5\% & 5000  & 10    &       &       &  \\
    \hline
    \multicolumn{6}{l}{\textbf{customized segmentation}} \\
    \hline
    100   & 500   & \multirow{4}[2]{*}{5} & \multirow{4}[2]{*}{$1\times10^{-2}$} & \multicolumn{1}{l|}{Adam:} & \multicolumn{1}{l}{w/ data aug: } \\
    50    & 1000  &       &       & $\beta_1=0.9$, & \multicolumn{1}{l}{random mirror, } \\
    25    & 1600  &       &       & \multicolumn{1}{l|}{$\beta_2=0.999$} & \multicolumn{1}{l}{random affine} \\
    10    & 2000  &       &       &       &  \\
    \hline
    \multicolumn{6}{l}{\textbf{customized stylization}} \\
    \hline
    50    & 400   & \multirow{4}[2]{*}{5} & \multirow{4}[2]{*}{$1\times10^{-4}$} & \multicolumn{1}{l|}{Adam:} & \multicolumn{1}{l}{w/ data aug: } \\
    25    & 400   &       &       & $\beta_1=0.5$, & \multicolumn{1}{l}{random mirror, } \\
    15    & 500   &       &       & \multicolumn{1}{l|}{$\beta_2=0.999$} & \multicolumn{1}{l}{random affine} \\
    10    & 600   &       &       &       &  \\
    \hline
    \multicolumn{6}{l}{\textbf{shadow removal} (w/ `clean images')} \\
    \hline
    2500  & \multirow{4}[2]{*}{300} & \multirow{4}[2]{*}{25} & \multirow{4}[2]{*}{$1\times10^{-4}$} & \multicolumn{1}{l|}{Adam:} & \multicolumn{1}{l}{w/ data aug: } \\
    500   &       &       &       & $\beta_1=0.5$, & \multicolumn{1}{l}{random mirror, } \\
    100   &       &       &       & \multicolumn{1}{l|}{$\beta_2=0.999$} & \multicolumn{1}{l}{random affine} \\
    50    &       &       &       &       &  \\
    \hline
    \end{tabular}%
  \label{tab:supp_training_details}%
\end{table*}%

\vspace{4mm}
\noindent\textbf{Face Semantic Segmentation}

All quantitative evaluations are conducted on the Helen dataset \cite{helen}.
The Helen dataset contains 2,330 face images of $400\times 400$ pixels with 11 manually labeled components. 
It has a training set with 2,000 images, a validation set with 230 images and a test set with 100 images. 
Note that the hair region is annotated through an automatic matting algorithm and is not accurate enough compared to other annotations. 
Following the common practice, the hair region is not considered during evaluations.

The images of customized segmentation data are randomly selected from Helen's training set.
We use 100, 50, 25 and 10 training images for ablation studies.
The customized annotation is the combination of newly defined regions (forehead, left/right cheeks, upper/lower and left/right eye sockets, under nose region, chin, eyeglass/frame, moustache and beard) and existing labels (left/right eyebrows, left/right eyes, nose, upper/lower lips, in mouth region) from the Helen dataset.
We annotate these images using the LabelMe toolbox with polygons.

The evaluation metric for quantitative studies is the F1 score.
We report the score for each category.
Following the common practice in all previous works, the overall F1 score is the macro averaged score of facial components, excluding face skin, hair and background.

\vspace{4mm}
\noindent\textbf{Facial Image Stylization}

The customized artistic styles are generated on the randomly selected images from CelebA-HQ dataset \cite{celebahq}.
We collect 50 images for both styles and use all, 25, 15 and 10 pairwise data for training.

\vspace{4mm}
\noindent\textbf{Facial Shadow Removal}

For facial shadow removal task, we use synthesized shadow images for training following the pipeline of \cite{face_shadow_removal}.
First, we collect images that have no foreign shadows and extreme illumination conditions, namely the `clean images'.
Then portrait areas are cropped out using a face/human segmentation model.
To generate training samples, a random mask is applied on the portrait area.
Pixels inside the random mask are randomly decayed to imitate a real shadow. 
The random mask can be generated either using the Perlin noise or an object silhouette.
More details can be found in \cite{face_shadow_removal}.
The model's training and inference are conducted on the portrait areas only.
The processed result is obtained by combining the model's output and the image's original background.
In the few-shot experiments, we collect 2,500, 500, 100, 50 and 5 `clean images'.
As a comparison, \cite{face_shadow_removal} used 5,000 `clean images'.
The number of synthesized training data is 5,000 despite the exact amount of `clean images'.

\section{Training Details of the Unsupervised Learning}
Based on the implementation of \cite{BrowatzkiW19}, the network backbone combines a standard ResNet18 \cite{resnet} as the encoder with an inverted ResNet18 as the decoder. 
Both encoder and decoder contain about 10 million parameters each.
The dimension of latent feature $z$ is 99.

The adversarial auto-encoder is firstly training for 50 epochs on $128\times128$ input and reconstruction resolutions with batch size as 100.
Then the network is further trained on the $256\times256$ resolution for another 50 epochs with batch size as 50.
The optimizer is Adam with learning rate as $2\times 10^{-5}$, $\beta_1=0$, $\beta_2=0.999$.
The data augmentations include random mirror, translation, rescaling and rotation.

\section{Training Details of the Supervised Model Adaptation}
The experimental settings of different applications are listed in Table \ref{tab:supp_training_details}.

\clearpage

\end{document}